# Personality Analysis for Social Media Users using Arabic language and its Effect on Sentiment Analysis

Mokhaiber Dandash
Electrical and Computer Engineering, artificial intelligence and robotics group, PhD Student, University of Tehran, Beirut, Lebanon, m.dandash@ut.ac.ir
0000-0001-7445-509X

Masoud Asadpour
Electrical and Computer Engineering, Assistant Prof. Director of Social Networks Lab, University of Tehran, Tehran, Iran, asadpour@ut.ac.ir
0000-0003-0299-4523

**Abstract**

Social media is heading towards more and more personalization, where individuals reveal their beliefs, interests, habits, and activities, simply offering glimpses into their personality traits. This study, explores the correlation between the use of Arabic language on twitter, personality traits and its impact on sentiment analysis. We indicated the personality traits of users based on the information extracted from their profile activities, and the content of their tweets. Our analysis incorporated linguistic features, profile statistics (including gender, age, bio, etc.), as well as additional features like emoticons. To obtain personality data, we crawled the timelines and profiles of users who took the 16personalities test in Arabic on 16personalities.com. Our dataset, "AraPers", comprised 3,250 users who shared their personality results on twitter. We implemented various machine learning techniques, to reveal personality traits and developed a dedicated model for this purpose, achieving a 74.86% accuracy rate with BERT, analysis of this dataset proved that linguistic features, profile features and derived model can be used to differentiate between different personality traits. Furthermore, our findings demonstrated that personality affect sentiment in social media. This research contributes to the ongoing efforts in developing robust understanding of the relation between human behaviour on social media and personality features for real-world applications, such as political discourse analysis, and public opinion tracking.

## 1 INTRODUCTION

In our previous work [1], we outlined the need for an improved dataset and a broader spectrum of features for analyzing personality traits using social media data. Building on these future directions, the current study aims to enhance the dataset and expand the feature set to provide a more comprehensive analysis. By leveraging a substantial dataset named "AraPers", we focus on extracting features for personality analysis and exploring the intricate connection between sentiment and personality.

Users in social networks reveal their personality characteristics, the study of these characteristics is important for several fields, such as social science, psychology, marketing, and others. People on social networks typically demonstrate their real nature [2] and leave traces in their online behavior, which mirror their actual personality [3]. The term personality refers to characteristic patterns of behaviors, thoughts, and emotional responses that evolve from genetic and environmental aspects and form relatively reliable individual differences [4].

Social media platforms are among the most frequently visited websites online [5], and they are the primary way people around the world connect to their digital environment [6]. The collaborative nature of these platforms has led to an increasing interest in investigating the role of personality in anticipating behaviors on social media [7–11].

Personality traits are reflected in people's communication practices and behavior, and thus linked to forms of media usage [12–15]. Given the significance of social media, many research studies have been dedicated to this area. However, most of these studies have concentrated on English and other European languages. Very little studies have actually analysed morphologically rich languages such as Arabic, despite the increasing number of Arabic internet users and the growing volume of online content in Arabic.

Twitter reveals many aspects of the user's life, either through their account profiles or through their usage of the platform. One of these aspects is personality. The main question is whether social media profiles can contribute to identifying personality, so that we can relate the information revealed by the social media contributors to their personality traits.

To further study personality effect there are several challenges to overcome, most notably:

- Complexity and ambiguity of the Arabic language.
- Lack of annotated dataset for this topic.
- Obtaining personality trait tests for our case study.
- Finding the most suitable ways to gather data from case study participants.
- Ensuring the presence of participants on social media platforms.
- Creating ways to enhance participation and promote the study.
- Dealing with multi-class topics.
- Finding efficient ways to analyze and produce best possible model.

To address these concerns, we focused on obtaining data from individuals who took the Arabic personality test (https://www.16personalities.com/ar), and shared their results on twitter. Therefore, we crawled participant's profiles and timeline, and we got an impressively large dataset. We divided this dataset into categories, to examine it in a more detailed manner.

We applied different statistical tests on user profiles, tweets and interactions (retweets, reply) to link these statistics back to personality types to gain knowledge. Additionally, after dividing the data into an equal number of tweets per personality, we applied different machine learning algorithms to develop models for this aspect. Finally, we investigated how personality affects sentiment analysis by analysing sentiments associated with each personality type and connecting them to personality features.

The structure of this paper consists of four main sections:
Section 2 provides a background on the personality detection in social media, highlighting the challenges associated with Arabic language processing. Section 3 explains our research methodology. Section 4 discuss our findings, and lastly, Section 5 presents the conclusion.

## 2 BACKGROUND

Recent years have witnessed growing research interest, in how personality features are derived from the traces that people leave on social media platforms. Personality traits are frequently revealed through self-expression and activities in social networks [3, 16]. Recent work connecting personality traits with language use on social networks has carried a number of perceptions regarding linguistics, grammar and syntactic features [17]. For example, people who have higher score on openness, show themselves more often with emotion and expressive language.

Social media is embodiment of behavioural patterns [18]. Extracted features related to language, sentiment, and behavioural patterns can reveal individuals' personality, which can be used as a representative for personality prediction [18, 19, 20, 21];

Several studies have predicted personality traits in social networks [22, 23, 24, 25, 26, 27, 28, 29, 30, and 31]. For example, language features were used by [31] to predict personality traits on social network environments, while [23, 24 and 26] concentrated on the info contained in the text of the tweets that mirror personality.

A different approach was taken by [22, 28], who used users' online actions to predict personality traits. In [28] personality and twitter use were studied, including users' number of followers. The study showed that publicly available data, such as the number of statuses, number of followers, and number of reflecting showing user behaviour, can be used to predict personality.

Moreover, [29] employed a different approach based on acquaintances' evaluations of individuals in order to make a personality classifier, established on behavioural signs from Facebook profiles. Both language and behaviour features have been used by [25, 27, 32, 30].

Deep learning architectures have been applied in the methods of [30, 33, and 34] to address personality prediction difficulties for Facebook user data by feeding different text, behavioural, or sentiment features into artificial neural networks.

### 2.1 Difficulties in Arabic Processing

Arabic language is one of the dominant languages worldwide; it is one of the six official languages of the United Nations (UN), and more than 400 million people speak it. It has three main forms: classical Arabic (the language of the Quran, Islam's holy book), modern standard Arabic (MSA), and dialectical Arabic.

Arabic language is written from right to left, and is deprived of upper or lower cases. In addition, diacritics are most often omitted, the Arabic handwriting uses pronunciation marks as short vowels. These are located either above or below the letters to provide the correct pronunciation and enrich the meaning of the word.

Arabic language has a very complex and rich morphology, a word in Arabic exposes several morphological aspects such as derived or inflectional morphology. In Arabic, words change according to categories like tense (past and present), person (1st, 2nd and 3rd), number (singular, dual and plural), and gender (feminine and masculine).

Arabic is also agglutinative, which means that the word may be joined with a set of affixes. The complexity of the Arabic word structure is one of the main difficulties that researchers face when dealing with Arabic analysis.

This complexity needs the development of suitable systems that are able to work with tokenization, stemming, lemmatization, and POS tagging. The main purpose of morphological analysis is to divide words into morphemes and to tie up each morpheme with a morphological information such as stem, root, part of speech (POS), and affix.

Nowadays, many morphological analysers for Arabic are already developed; some of these are freely available while the rest have a profitable purpose. However, these systems suffer from important limitations especially in handling ambiguity that can result from the elimination of diacritics (short vowels which are necessary for the meaning, they are often excluded as Arab readers can usually deduce them from context), and the free word-order nature of Arabic sentence.

For communication purposes, Arabic speakers usually use informal Arabic rather than MSA. There are around 30 major Arabic dialects that vary from MSA and from each other phonologically, morphologically, and lexically (Habash, 2010) [35]. Furthermore, Arabic dialects have no standard orthographies and no language colleges; therefore, using tools and resources designed for MSA to process Arabic dialects produces clearly low performance.

## 2.2 Personality Test

There has been little work done on personality analysis in the Arabic language using social networks as a data source. However, there are plenty of works done in English language such as [36-41]. They mainly concentrate on the Myers-Briggs type indicator (MBTI) or big five (BF) traits, and rely on twitter GNIP API and the "My Personality" Facebook application dataset, which reveals personality based on an online questionnaire.

Other studies have explored personality analysis using profile pictures [37] and other engagement features rather than relying solely on textual data from social networks. To proceed with our research, we should first provide an overview for the most known personality test types.

- ❖ The Big Five (BF):

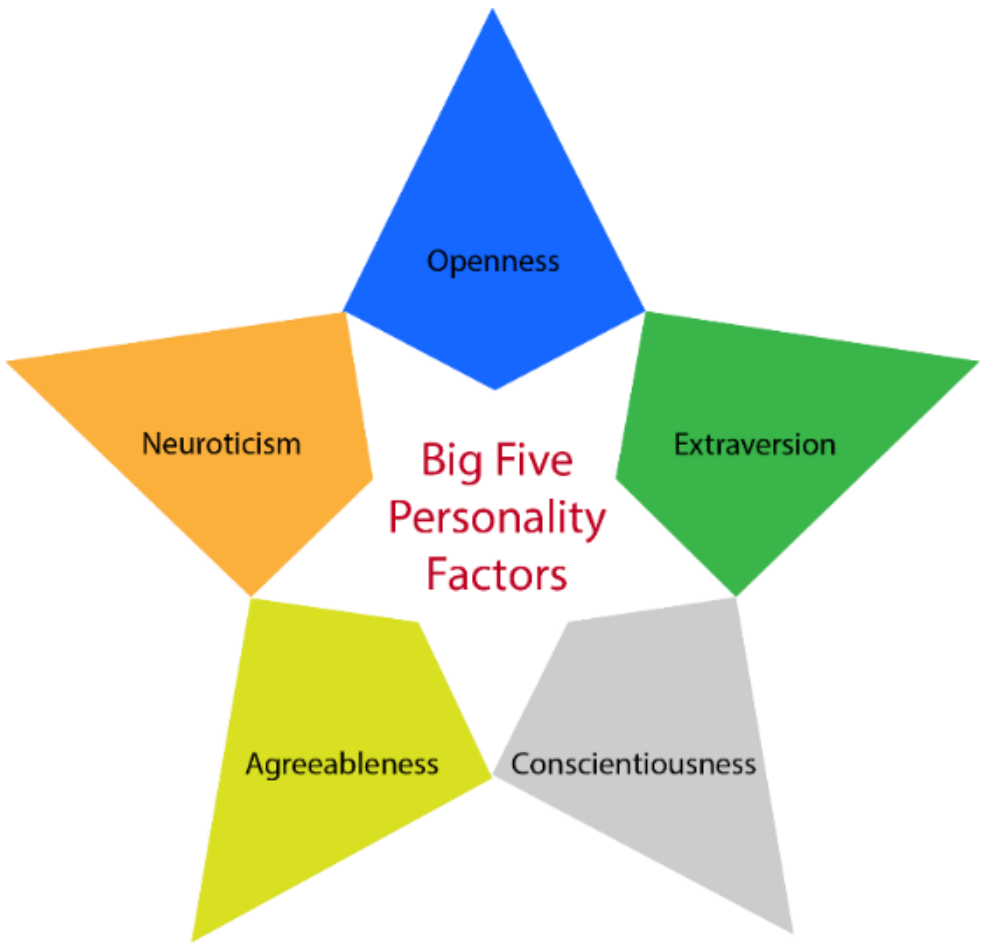

Figure 1: Big Five Personality Factors [42]

Big five (Fig. 1), also known as the five-factor model (FFM) and the OCEAN model, is a classification method for personality traits. The five factors are:

-Openness to experience: reflects degree of curiosity, creativity and novelty. People in this category are independent and open-minded.

-Conscientiousness: individuals in this category are organized and dependable, and prefer planned rather than spontaneous behaviour.

-Extraversion: people in this category are energetic, attention seeking, talkative, and tend to expresses their ideas.

-Agreeableness: individuals tend to be compassionate and cooperative rather than suspicious, and non-argumentative.

-Neuroticism: individuals in this category are prone to stress, and express unpleasant emotions like anger, and anxiety.

- ❖ Myers–Briggs Type Indicator (MBTI):

-According to [43-45] Myers–Briggs type indicator (MBTI) is *"An introspective self-report questionnaire indicating different psychological preferences in how people perceive the world around them and make decisions."*

-With the concept [46] posits that: *"MBTI has speculated that humans experience the world using four principal psychological functions – sensation, intuition, feeling, and thinking – and that one of these four functions is dominant for a person most of the time".*

The MBTI sorts some of these psychological differences into four opposite pairs, with a resulting 16 possible psychological types.

The four opposite pairs are:

1. Extraversion (E) vs. Introversion (I)
2. Sensing (S) vs. Intuition (N)
3. Thinking (T) vs. Feeling (F)
4. Judgment (J) vs. Perception (P)

For instance:

- ESTJ means extraversion (E), sensing (S), thinking (T), judgment (J)
- INFP means introversion (I), intuition (N), feeling (F), perception (P)

The 16 types typically referred to by an abbreviation of four letter i.e. the initial letters of each of their four type preferences (except in the case of intuition, which uses the abbreviation "N" to distinguish it from introversion).

## 3 METHODOLOGY

Social media outlets nowadays are tribunes for individuals to freely express their thoughts, feelings, attitudes, and support on various subjects affecting communities. They act as windows to the broader world, allowing every person to positively or negatively influence and interact with other societies, fellow citizens, and fellow humans, all on a topic-by-topic basis.

Social media platforms offer users opportunities to engage, regardless of geographical distances, through actions like liking, sharing, posting videos, and more. The diverse personalities of users play a crucial role in shaping the varied content and interactions observed across social media.

Consider contemplating the impact that personality might have on shaping various aspects of social media data, such as the number of likes, shares, retweets, and the overall abundance of posts and tweets.

Social media analysis is not a standalone concept, from our perspective; it requires assisting algorithms to extract knowledge from it, in a human-understandable structure. We aim to enhance these assisting algorithms in this paper.

As one strives to attain proficiency in data analysis, various challenges emerge, including issues related to data collection, adherence to analysis standards, achieving meaningful data interpretations, and determining the appropriate type of data to extract from social media outlets, among others.

Up to our knowledge, there is no substantial dataset available for personality analysis in Arabic language, apart from the dataset created in our previous work [1]. To add, collecting personality-related datasets has become even more challenging, especially after the Cambridge Analytica crisis [47]. To solve this dilemma, we deployed a new method to collect data, using the 16personalities test for users who share their results on twitter (fig.2).

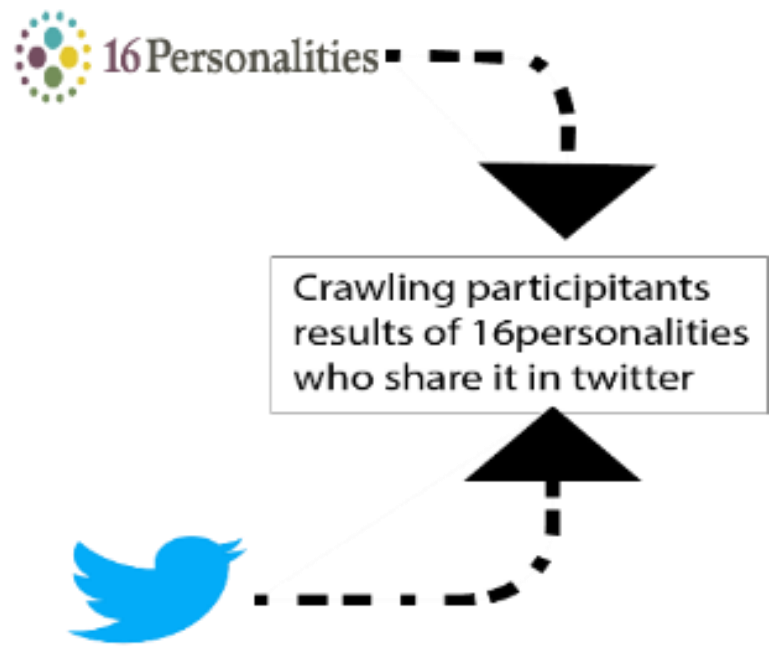

Figure 2: 16personalities result shared on twitter

To investigate the different aspects and relations, between user personality traits, sentiment and the way of expression on social media outlets, we followed a series of steps and milestones as follow:

1) Observation: We observed how the “16personalities” website shares results on twitter for applicants, noting that they are shared in a specific format (e.g., “القائد” شخصية https://16personalities.com/ar/شخصية-entj #16Personalities via @16Personalities).
2) Search: We used Twitter’s advanced search option to find tweets containing keywords related to personality traits (e.g., البطل، المنطقي), mentioning (@16personality), and using the hashtag (#16Personalities) with the language set to (Arabic).
3) Data Crawling: We crawled the data using the “Selenium” package in python with “XPath” as navigation technique for html tags, to extract user information.
4) Data Collection: We collected the timeline and profile features for the users through twitter API and other indirect techniques, with related personality types as labels.
5) Data Categorization: The data was divided into three categories for different statistical experiments:
   a. Profiles includes bio, followers, friends, Likes, statuses, verified, gender, location, age, zodiac and media.
   b. Tweets includes tweets and quoted tweets written by users, referred to as “level zero”.
   c. Interactions includes replies and retweets by users, referred to as “level one”.
6) Statistical Analysis: We conducted statistical analysis on the data, focusing on account statistics (e.g. count of media, gender), bio features (e.g. count of emoji), and features from “level zero” and “level one” (e.g. count of mentions).
7) Feature Conversion: The data from “level zero” was distributed equally across the 16 personalities, and converted into a vector composed of quantitative and categorical features using different methods.
8) Machine Learning: We applied machine learning and neural network techniques to relate the features extracted from twitter to personality types.
9) Sentiment Analysis: We performed sentiment analysis on each personality dataset in “level zero” to determine the percentage of the positive, negative and neutral tweets for each personality type.
10) Effect Examination: We examined the effect of personality on sentiment analysis.

### 3.1 Experimental Setup

Users interact more with each other, when there are similarities in their personalities; as the saying goes “Similarity breeds connection” [48]. We examine relationships between users based on actions performed on their social media accounts (e.g. retweet).

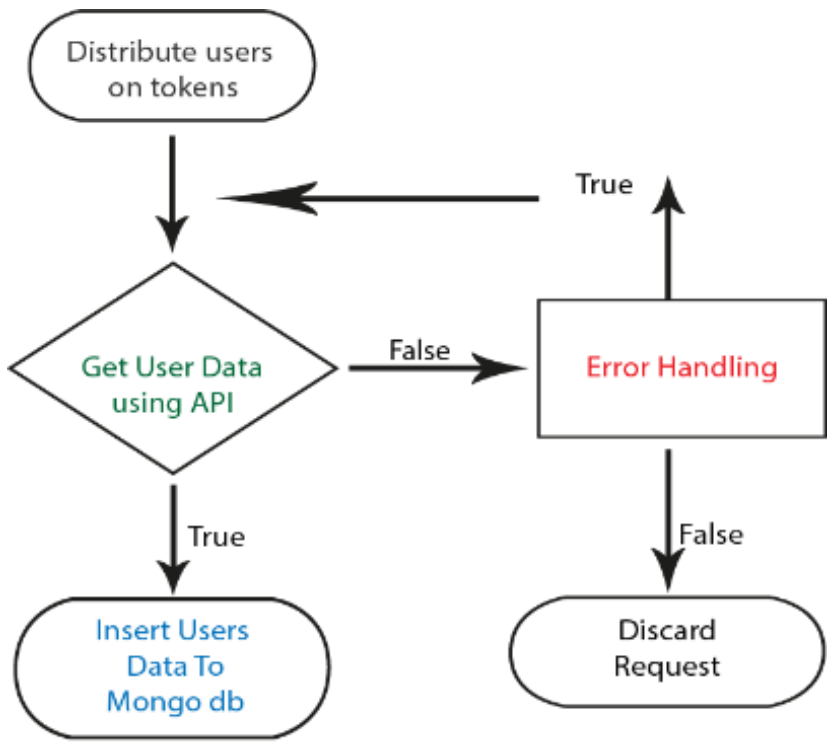

Figure 3: Data gathering flow chart

We gathered information from twitter, using the official API, as shown in the (fig.3), along with other methods described later. All the experiments implemented in python on a desktop computer with an Intel® Core™ i7-7700 CPU @ 3.60 GHz and 16GB of RAM. The GPU used was an Nvidia GeForce GTX 1060 6GB, and we also used google cloud and google Colab.

### 3.2 Subjects of the experiment

The individuals included in our research are from different countries, ages, genders, places, educational backgrounds, all of whom are Arabic speakers. Gathering the necessary dataset for personality analysis is fundamental step, as the dataset is crucial to our work.

Using Twitter's advanced search, we collected 5,509 tweets from users declaring their personality. We employed web crawling to obtain this data and, after eliminating noise, we identified 3,250 users with mentioned personality types, which we used as our case study. We retrieved the time line and profile for each user using twitter API.

We collected the accounts data (description and statistics) and approximately five million statuses. We divided the work into three categories (fig.4); profiles: we extracted user profile data and concentrated on features related to them such as number of followers, level zero: we collected 2,380,188 text (tweets: 1,898,064 and quoted tweets: 482,124), and level one: we analysed interactions, focusing on 223,358 reply and 1,992,023 retweet. These categories and their features will be discussed in the next section.

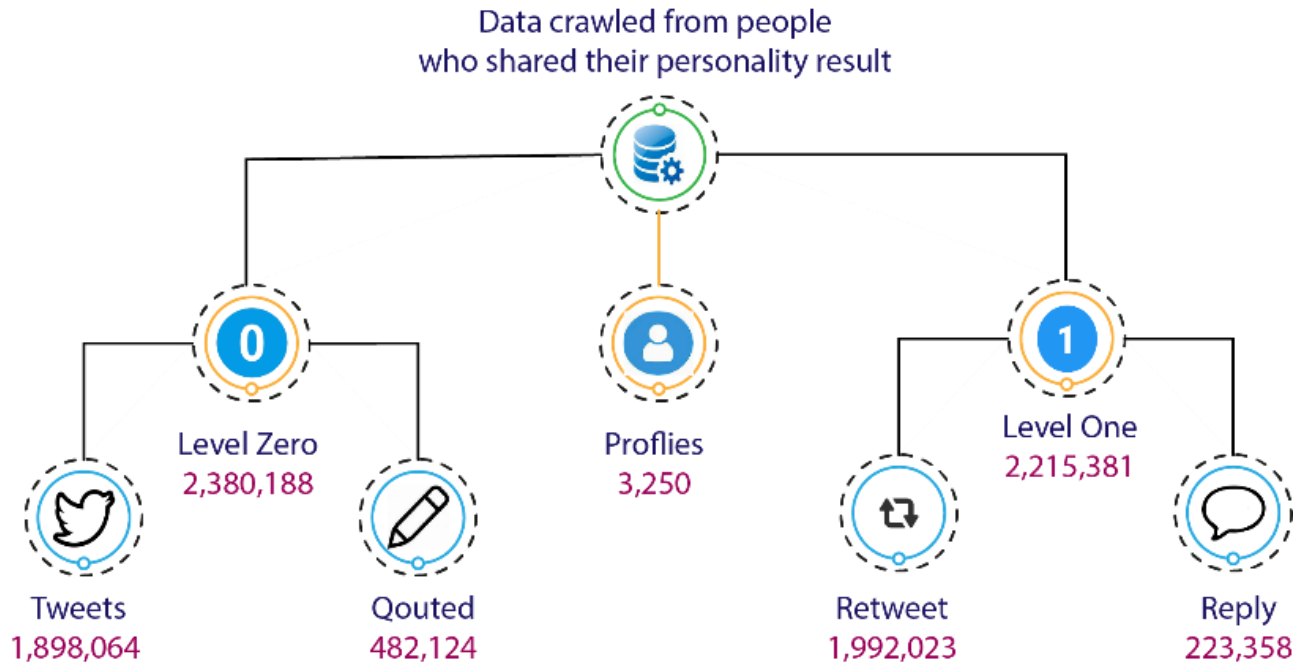

Figure 4: Data Distribution

We noticed that the number of tweets for each personality type in level zero varied significantly. To ensure unbiased analysis, we created an equally distributed dataset of tweets, named “Balanced Corpus”, containing 1088000 tweets, with 68000 tweets for each personality type.

### 3.3 Extracted Features

It is important to mention that after collecting the data we performed a normalization step so that we removed the stop words and all characters other than Arabic letters.

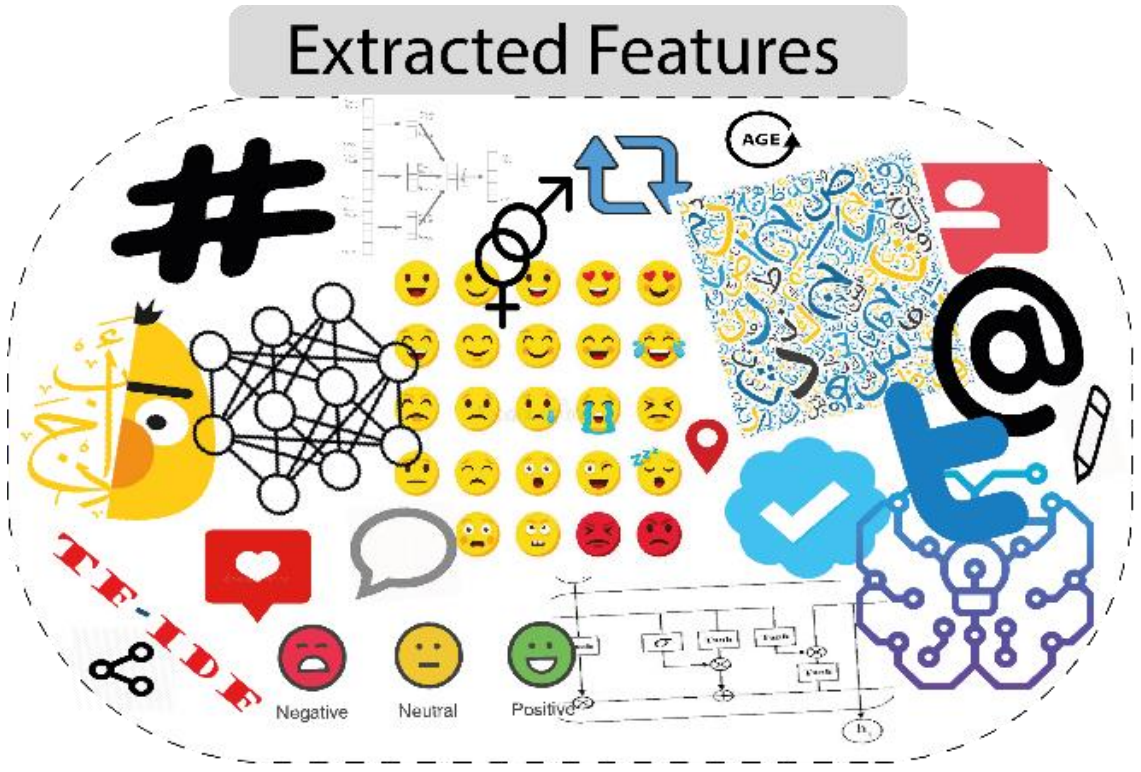

Figure 5: Features extracted

We used the following features (fig.5):

1) Profile statistics with respect to each personality: number of followers, friends, verified accounts, statuses, media, location, likes, age and zodiac all those statistics express user personality features.
2) Bag of words (Sivic et al., 2009) [49]: we denoted the existence of informative words (excluded some useless stop-words from a list) by assigning a column to that word and coded the column corresponding to that word with one (1) and set the other entries to zero (0). For every text, the entries of this vector has one in the columns that corresponds to the words that exist in the text and has zero for the other entries.
3) Term frequency inverse document frequency (TF-IDF) (Rajaraman et al., 2011) [50]: This is the same as the previous vector except that instead of ones, the TF-IDF score of the word (which is a real number) is used.
4) Word2vec (Mikolov et al., 2013) [51]: the texts are given to a neural network that transfers the words into a 300-axes space to use in deep learning. The network encodes the proximity of the words that appear near to each other in the text and for each word it gives a vector that uniquely specifies it in that space.
5) Multi-linguality: some contributors posted in English, which initially seemed undesirable, since our focus is on Arabic language, yet we used this feature to strengthen the results for some personality types, as the use of foreign language, might reveal additional aspects of their personality traits.
6) Gender: Categorical data related to gender.
7) Hashtags and Mentions: As used by (Quercia et al., 2011) [27], and (Golbeck et al., 2011) [52], we included the number of hashtags and mentions to explore their influence on personality.
8) Number of words: The total number of words used by users of every personality trait.
9) Emoji Usage: The count of emojis used by users of each personality type.
10) Word Density: Calculated as the number of characters divided by the word count.
11) Punctuation Characters: The number of punctuation characters used.

12) Number of Title Words: The count of words starting with an uppercase letter.
13) Number of Characters: The total number of characters in the text.
14) Effect of personality traits on sentiment analysis: explained in section (4.4).
15) Relation between language features and personality traits: This includes phrases and expressions frequently used that reflect specific personality traits, such as first person usage, expressions of compassion, love, opinion, happiness, and sadness. This will be explained in the next section.

# 4 RESULTS

To detect the relationship between personality and features, we can divide the results into different sections depending on the analysis perspective, as follow:

## 4.1 Profiles:

First, we concentrated on users' profiles, divided into five partitions: gender, statistics (e.g. number of followers), age or zodiac, location and bio.

### *4.1.1* Gender*:*

Twitter does not require users to specify their gender when registering, and there is no direct method to classify a twitter account as male or female. Therefore, we must rely on indirect techniques to determine gender.

Nonetheless, a common question arises about the gender distribution within various groups. Are our case study subjects male, female, or unknown? How can we programmatically determine the gender of a person on twitter?

The twitter API doesn't provide the gender information but there's a workaround. We extracted the profile picture URL of a twitter user, and employed AI algorithms to detect gender. For this mission we used Microsoft's face API due to its reliability and diverse features in this aspect.

Microsoft Azure face API is informative, well documented, and user-friendly option. It provides documentation with explanation through each phase of automating their API's prerequisites:
- Choose a language to use for automation (python)
- A URL address to access the image being analysed.
- A free subscription-key
- A location address provided on site.

We used face API to detect faces in an image using python code. The site offers snippets for integration; it sets up the python code that calls the face API. This section includes adding the subscription key which offers you access to the API code. The free subscription key is generated based on the chosen location.

Face API need the image to be at least 50 pixels in width and height, we resized the profile images to 400x400 pixels from its normal size 48x48 pixels that is provided by twitter API. (Twitter supports only normal and 400x400 pixels sizes for profile pictures).

The results can have different confidence levels, taking into consideration the profile picture condition (degraded, blur etc.). We obtained 2,228 gender detections categorized as woman, man, boy, or girl options. In some cases, where there was more than one person in the profile picture, the "people" option was used. Accounts without gender labels often used profile pictures such as text, icons, photos, drawings, null etc.

Some users even used pictures of their children as profile images, which affected the accuracy of gender distribution, as the profile images did not always correspond to the users' genders. Despite these issues, we achieved 90% accuracy after double-checking a random sample of 100 users by relating their first names to the predicted gender.

The percentages of gender detection and no gender detection for each personality trait are shown below:

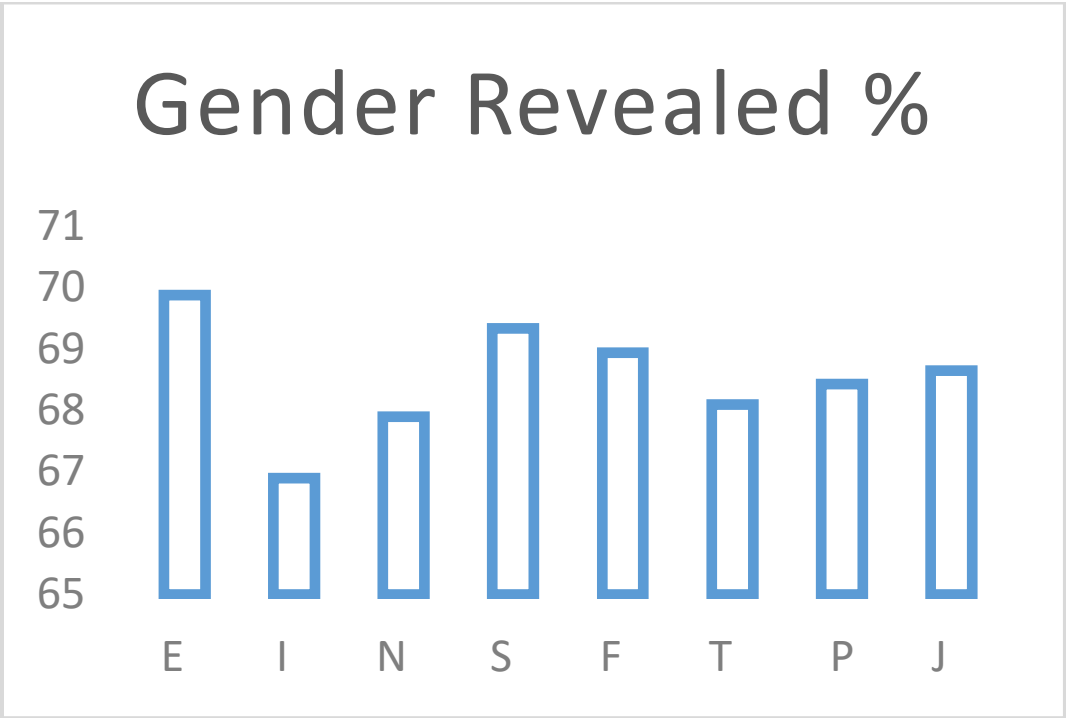

Figure 6: gender revealed percentage in each trait

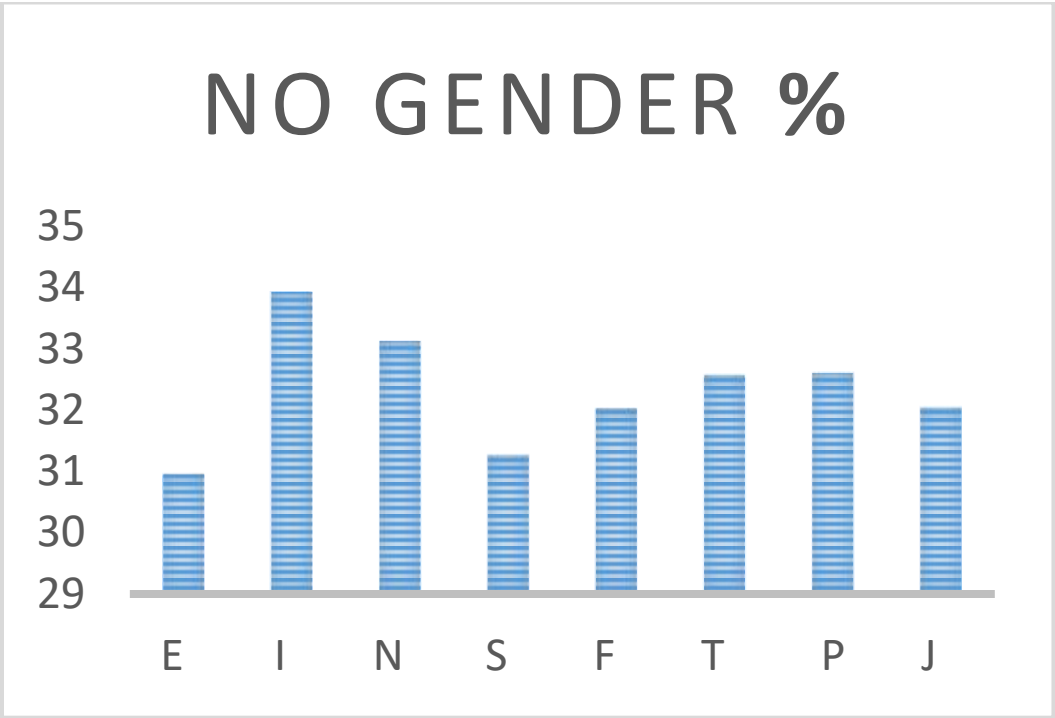

Figure 7: No gender revealed percentage in each trait

By examining these results, we can extract new features relating personality to gender revealing and willingness to post personal profile pictures. For instance, extroverted individuals are more likely to share their own profile picture compared to introverted individuals. Similarly, this trend is observed among sensing and feeling types (fig.6).

In case where users do not share any personal profile picture (fig.7), we can notice that introverted, intuitive and thinking individuals are more likely to do so. These observations plainly illustrate that personality traits manifest in social media behaviors.

Furthermore, when we categorize gender disclosures into male and female categories, we notice that females tend to exhibit more personality traits associated with caring and healing features (fig.8). Examples include INFJ (advocate, counselor), INFP (mediator, healer), and ENTP (debator, visionary).

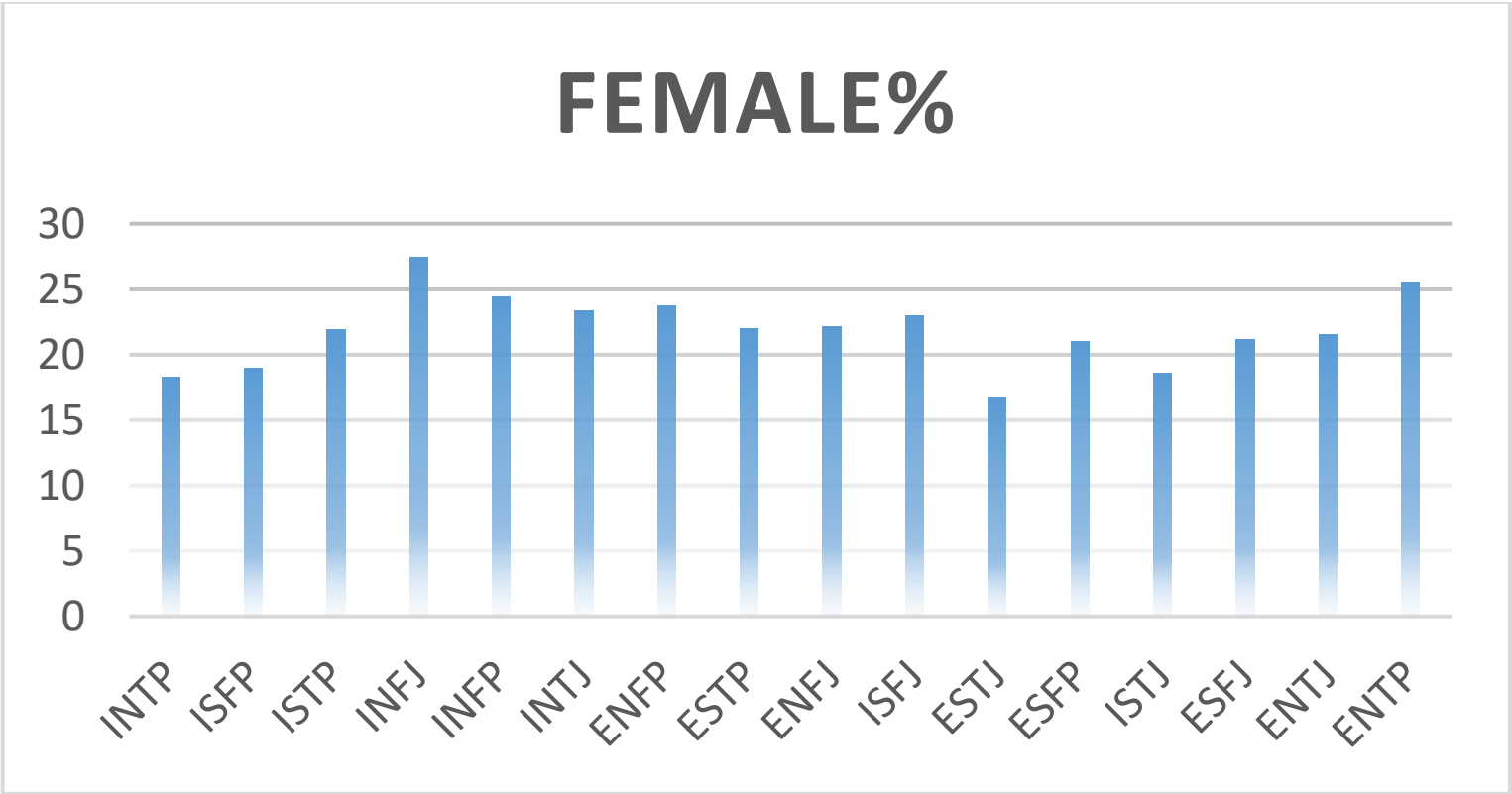

Figure 8: Female user's percentage in each personality

*4.1.2 Profiles statistics:*

We obtained the following results for profile statistics: (we used the average number as an indicator since the number of users per personality differs).

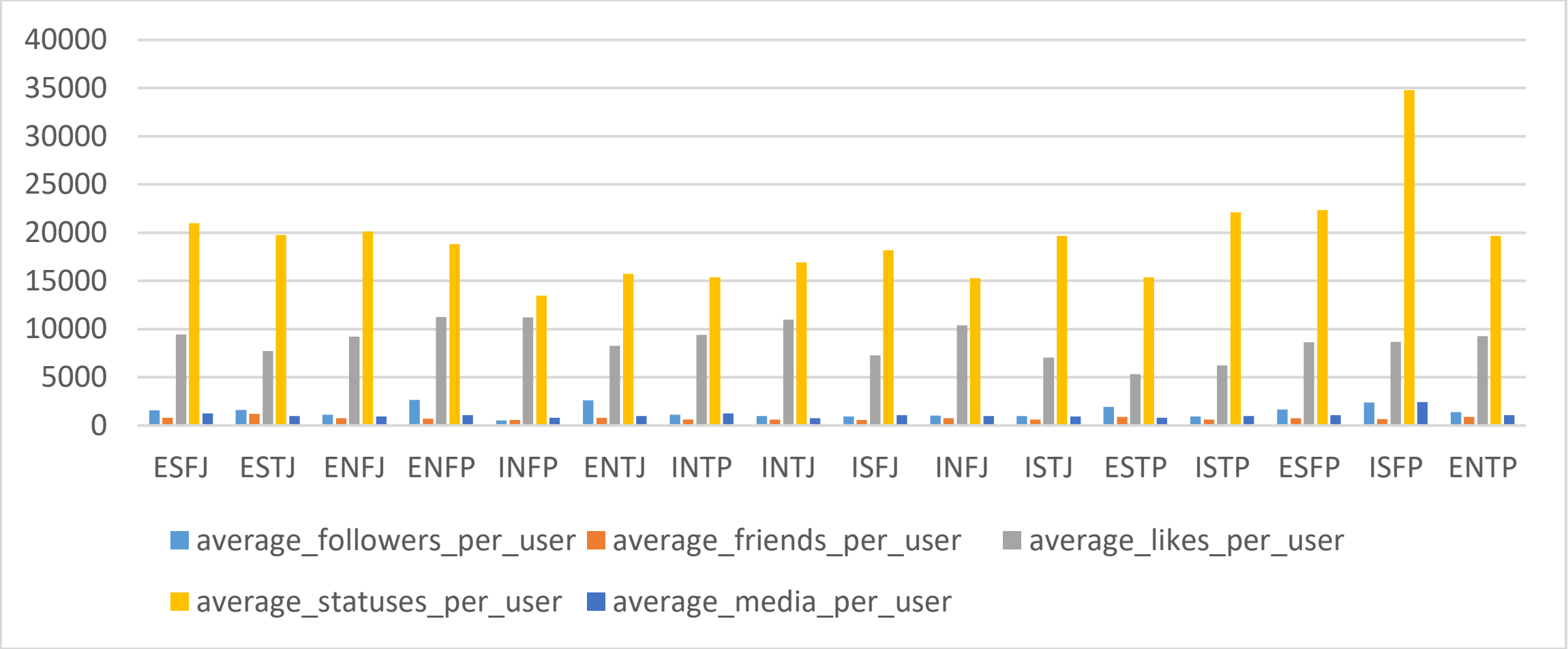

Figure 9: Average profile statistics per user with respect to personality

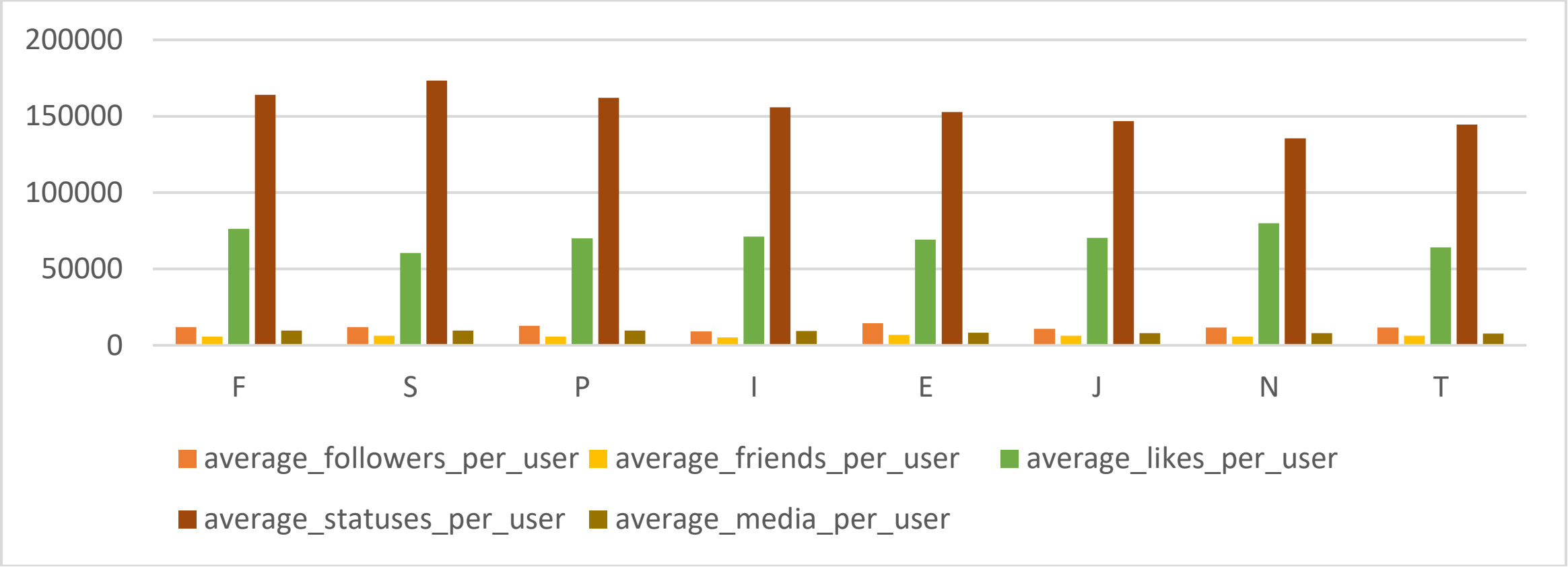

Figure 10: Average profile statistics per user with respect to personality traits

As shown in (fig.9) we can conclude that the ENFP (a diplomat personality type) and the ENTJ (an analyst personality type) have the most followers. ENFPs (campaigners) are “enthusiastic, creative and sociable free spirits, who always find a reason to smile” [53], and ENTJs (commanders) are “bold, imaginative and strong-willed leaders, always finding a way – or making one” [53].

The ESTJ (a sentinel personality type) and The ENTP (an analyst personality type) have the most friends:

ESTJs (executives) are “excellent administrators, unsurpassed at managing things or people” [53], and ENTPs (debaters) are “smart and curious thinkers who cannot resist an intellectual challenge” [53].

Explorers and sentinels have the most statuses and media (ISFP (adventurer) is “flexible and charming artists, always ready to explore and experience something new” [53], and ESFJ (consul) is “extraordinarily caring, social and popular people, always eager to help” [53].

Diplomats have the most Likes (ENFP (campaigner) and INFP (mediator) who is “poetic, kind and altruistic people, always eager to help a good cause” [53]).

On the basis of personality traits, as shown in (fig.10) we noticed that extrovert and perceiving have the most followers count, extrovert and thinking have the most friends count, intuitive and feeling people have the most likes, feeling and sensing people have the most statuses and media.

From the data, we observed that individuals identified as extroverts tend to have the highest number of verified profiles. All the above information underscores the idea that personality traits manifest in social media behaviour, so that extroverted and thinking individuals are generally more open to making new connections and friends, also followers are tending to follow those who express themselves vividly and exhibit creativity, characteristics often associated with extroversion and perceptive personalities.

Furthermore, our findings indicate that intuitive and feeling individuals are more likely to express emotions, leading to higher engagement in terms of likes. Conversely, those with feeling and sensing traits tend to share their thoughts and experiences more frequently, resulting in a greater number of statuses and media posts.

#### *4.1.3 Age and zodiac:*

Age and zodiac signs are additional features related to personality, we used the Burb Suite application to get the needed web request to obtain this data, since twitter API does not provide it. Age is determined if a user includes the year in their profile, whereas the zodiac sign is derived if the user provides only the day and month without specifying the year. This feature reveals somehow willingness to give private information. We observed that judging, intuitive, feeling, and extrovert people are more likely to provide this information about themselves, reflecting their readiness to interact with the world and share personal details.

We also find that there are five visibility options for users who provide their date of birth (DOB): self, public, mutual follow, followers, and following. The same options for year visibility apply if the user chooses not to display the year part only (“year_visibility”). Notably, female users tend to hide the year visibility more often using the (Self) option, for example: {"DOB": {"day": 10,"month": 7,"visibility": "Public","year_visibility": "Self"}}

Summing up all the choices under “Revealing” option, as all those decisions are consequences of revealing the date of birth in the first place. We noticed that ESTJ (executive) individuals, who are excellent administrators, unsurpassed at managing things or people, have the highest level of disclosure among those who choose to share certain details about their DOB.

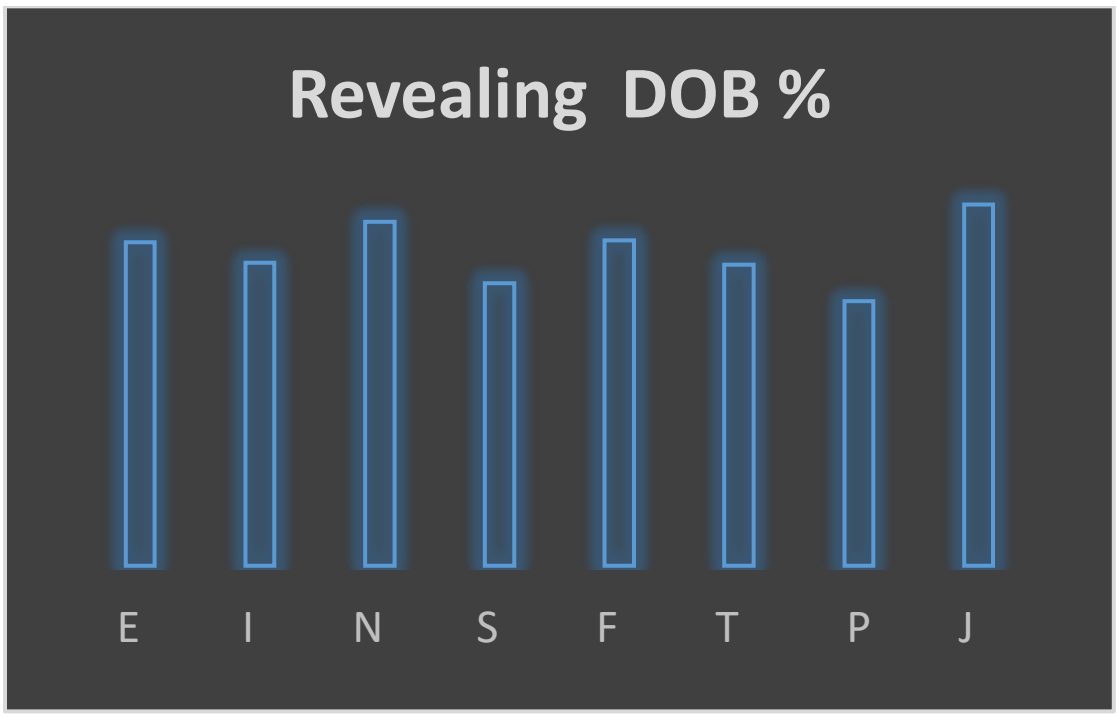

Figure 11: Revealing DOB percentage with respect to different personality traits

Also as shown in (fig.11), when comparing the results from the perspective of trait pairs, we noticed that extroverts reveal more than introverts, intuitive individuals reveal more than sensing individuals, feeling people reveal more than thinking individuals, and judging people reveal more than perceiving individuals. These observations align coherently with the characteristics of each trait with respect to its opposite.

#### *4.1.4* Location*:*

Users come from different countries around the world, including Saudi Arabia, Egypt, London Canada, morocco, Algeria, Oman, Kuwait, France, Sudan, turkey, Qatar, Portugal, Iraq, and many others (fig.12).

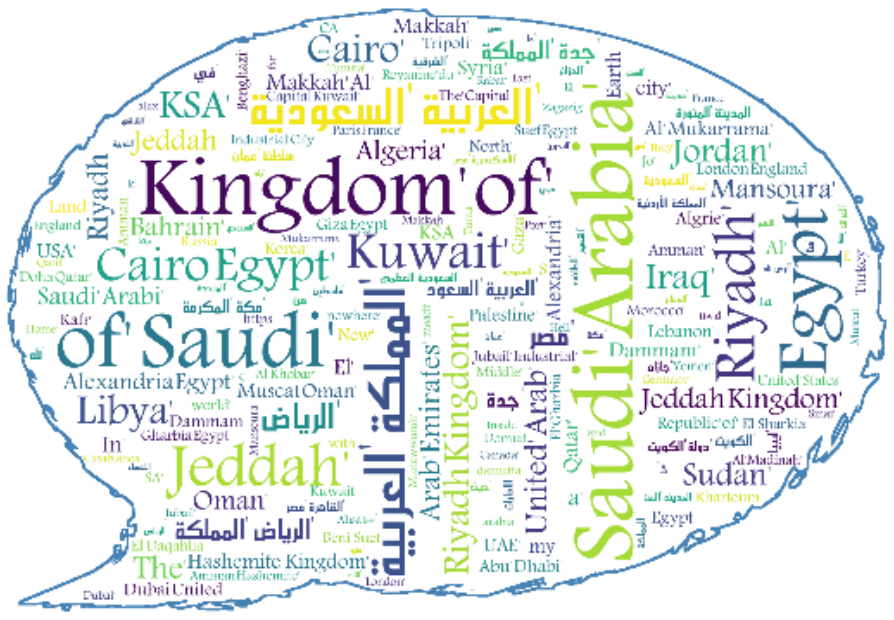

Figure 12: Different user's countries

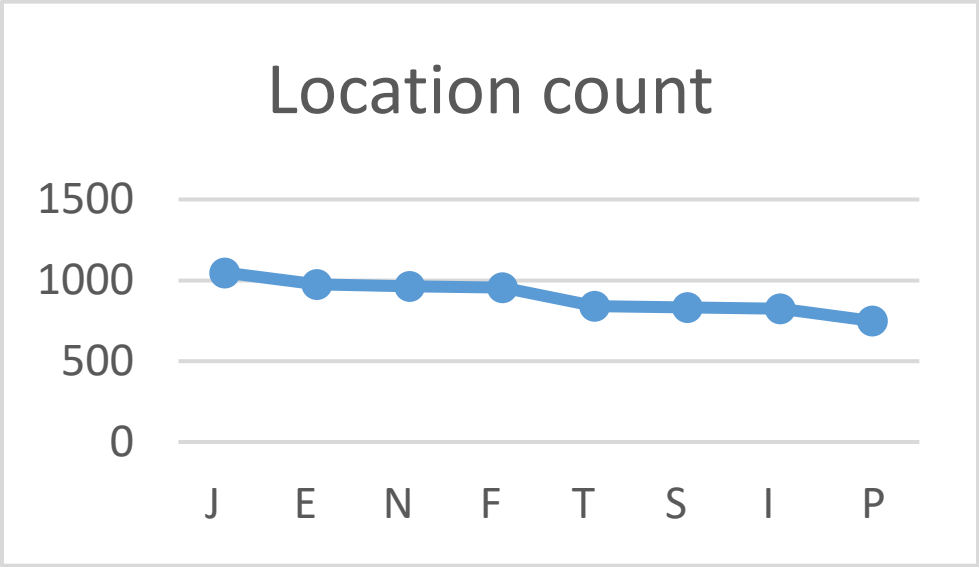

Figure 13: Location count with respect to personality traits

As shown in (fig.13), judging and extrovert individuals are more likely to announce their location. This behavior is consistent with the traits of being open and sticking to plan. In contrast, Introverted and perceiving individuals are more reserved and casual about revealing their location.

*4.1.5 Profiles Bio:*

We analysed the bio section of each user to study their characteristics in relation to personality traits, and results are summarized as follows:

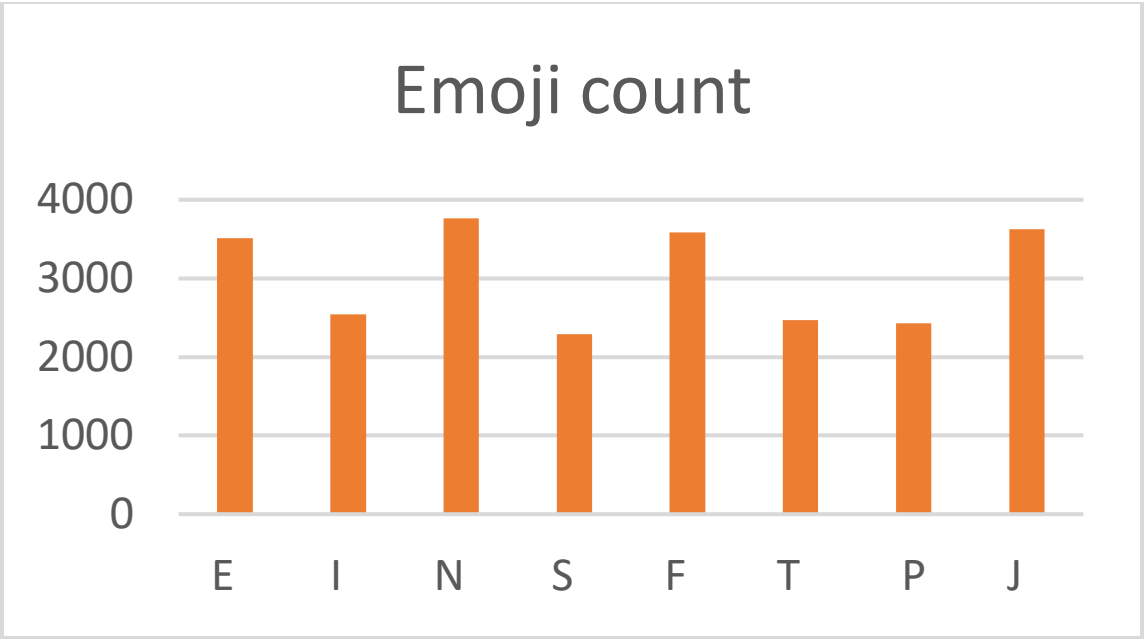

Figure 14: Bio Emoji count with respect to personality traits

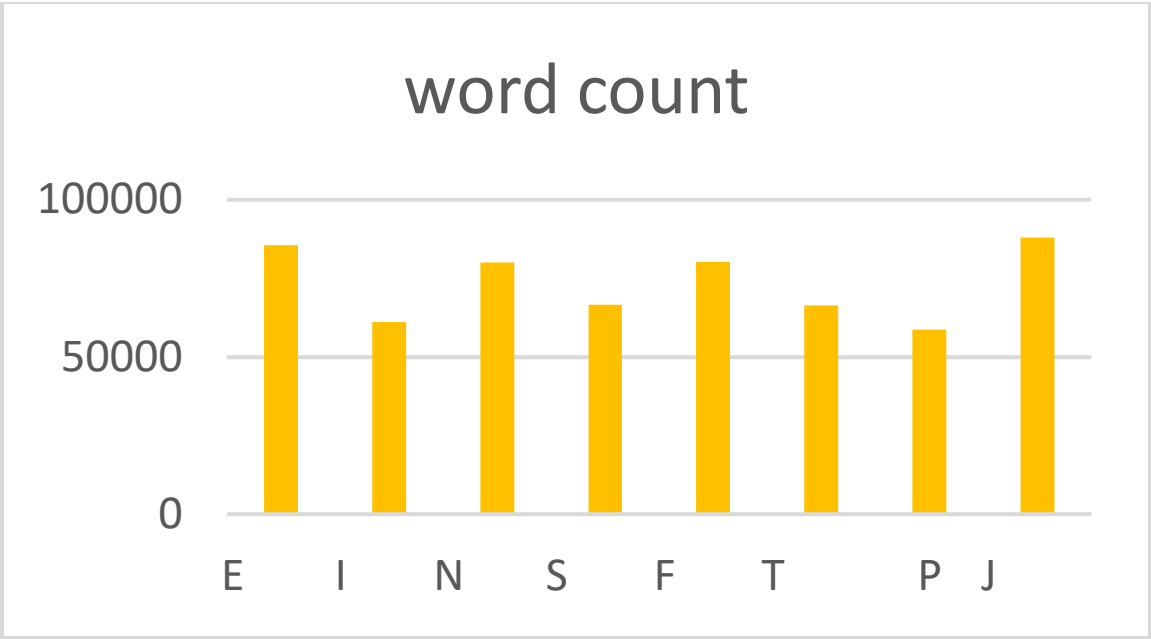

Figure 15: Bio word count with respect to personality traits

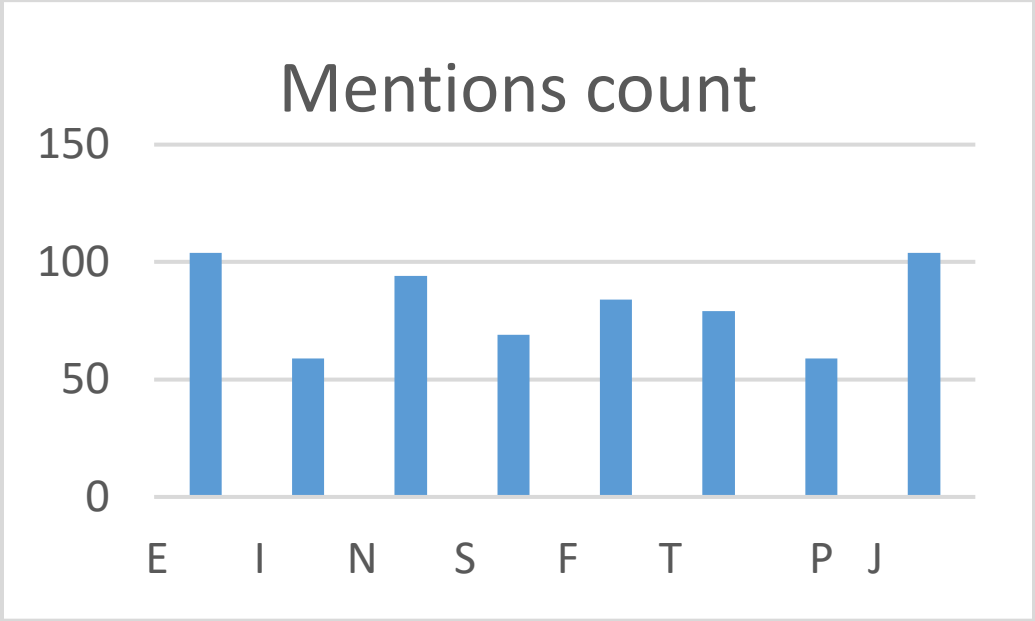

Figure 16: Bio mentions count with respect to personality traits

Figures 14,15 and 16 illustrate that intuitive, judging, and feeling individuals tend to use more emojis in their profiles. This pattern aligns with the expectations for these traits: intuitive individuals value creativity, judging individuals appreciate structure, and feeling individuals make decisions based on emotions.

Furthermore, extroverted individuals, who are typically more expressive and outspoken, along with judging and feeling personalities, use a greater number of words in their profile bios. Additionally, extroverted, intuitive, and judging individuals tend to include more mentions in their profiles. These findings affirm the distinctive characteristics associated with each of these personality traits. Our experiments also revealed that introverted individuals tend to use more English alphabet words.

## 4.2 Levels:

*4.2.1 Level zero:*

In this step, we examined tweets and quoted tweets from users, inspecting their data to obtain the following insights:
As depicted in Figures 17, 18, 19, and 20, we observed that individuals with judging, extroverted, and intuitive traits tend to use a higher number of hashtags, emojis, mentions, and words. It is noteworthy that similar patterns were observed in word density, the count of punctuation characters, title words, and character usage.

The Analyst personality type (commander), the Sentinel personality type (executive), and the Diplomat personality type (protagonist) use the most mentions, as these personalities tend to interact with people more. Sentinel types (consul and executive) and the Diplomat type (protagonist) use the most hashtags and words, reflecting their tendency to share opinions and advice. Diplomat types (protagonist, campaigner) and the Sentinel type (consul) use more emojis, highlighting their sociable and popular nature.

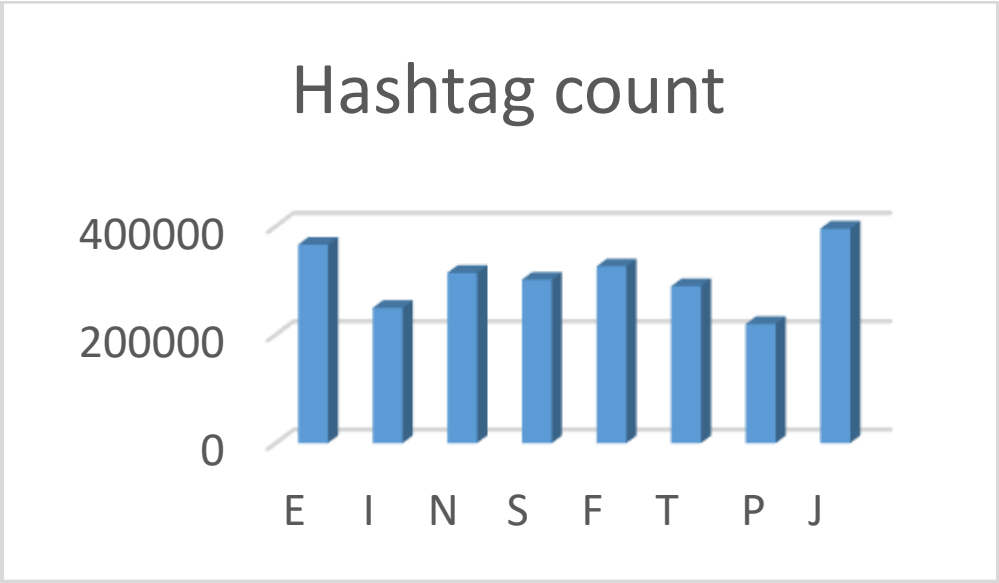

Figure 17: Hashtag count per personality trait

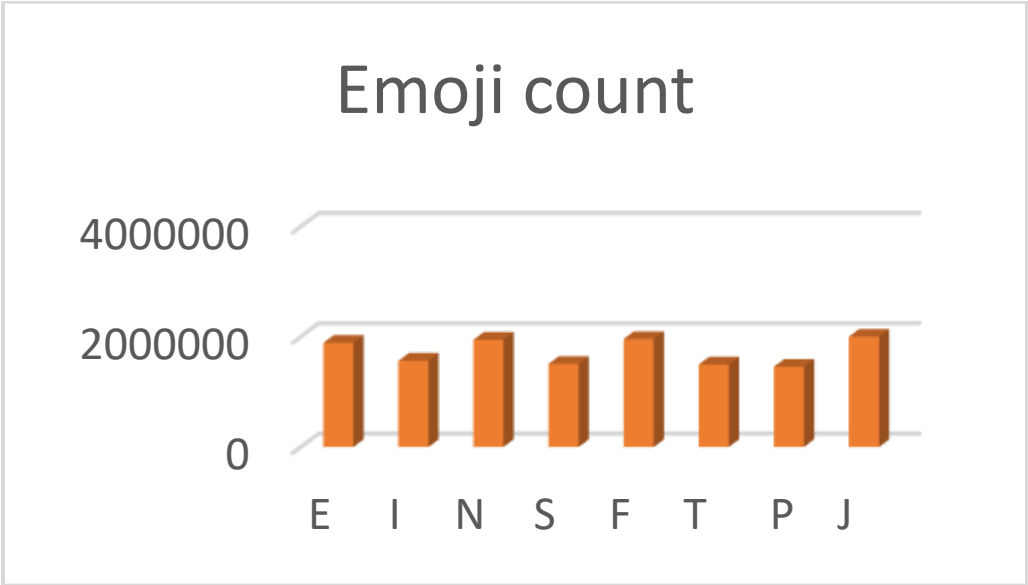

Figure 18: Emoji count per personality trait

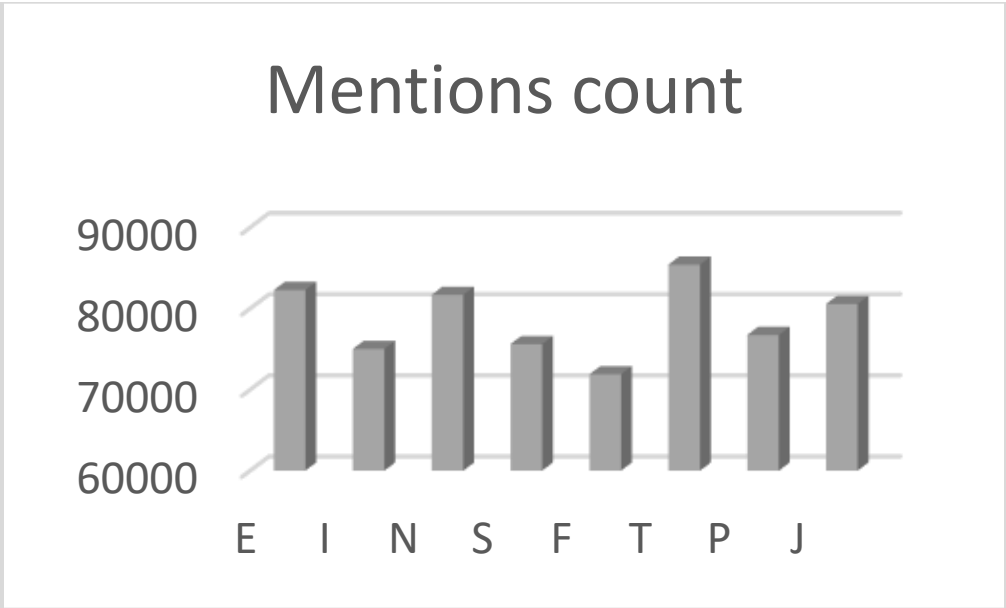

Figure 19: Mentions count per personality trait

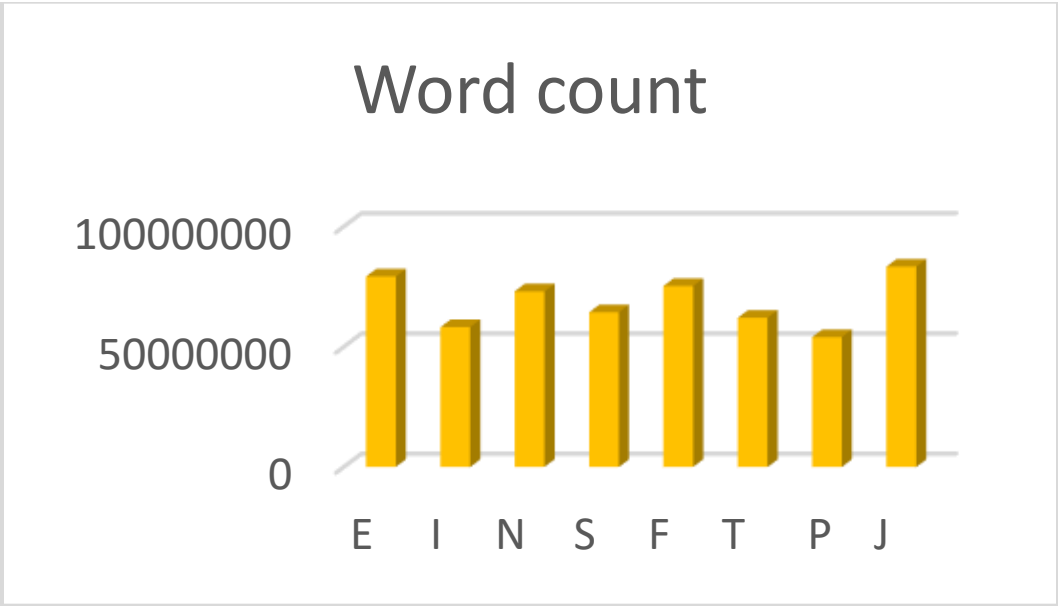

Figure 20: Word count per personality trait

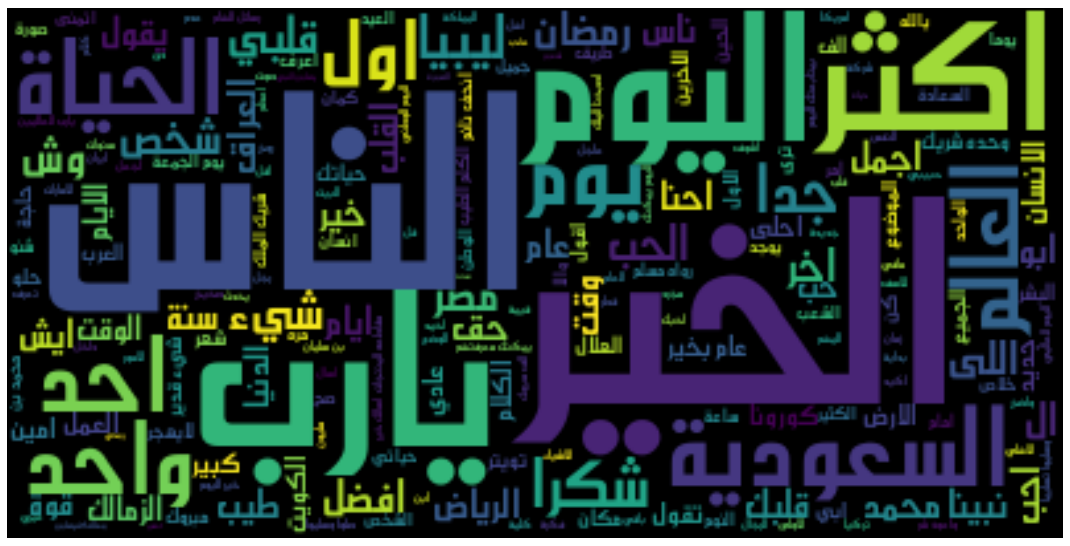

Figure 21: ESTJ word cloud

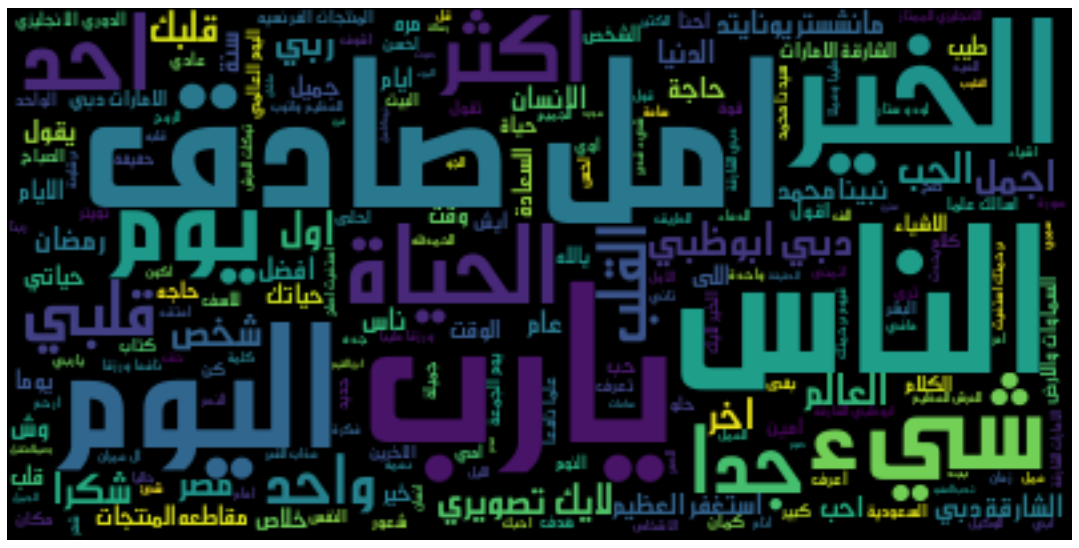

Figure 22: ISFJ word cloud

In (fig.21) and (fig.22), we observe that thinking and extroverted individuals often use specific words like (people, الناس) and (افضل, I prefer). In case of feeling and introverted individuals, they tend to use soft words such as “صادق” (honest) and “امل” (hope). This distinction provides valuable insights into the relation between phrases usage and personality traits.

### *4.2.2 Level one:*

Those patterns observed in “level zero” are also noticeable in interactions from “level one”. For instance, diplomats (e.g., protagonist and campaigner) and sentinels (e.g., consul and executive) use more emoji and words. Judging and extroverted individuals use higher number of words and emojis, as shown in (fig.23) and (fig.24).

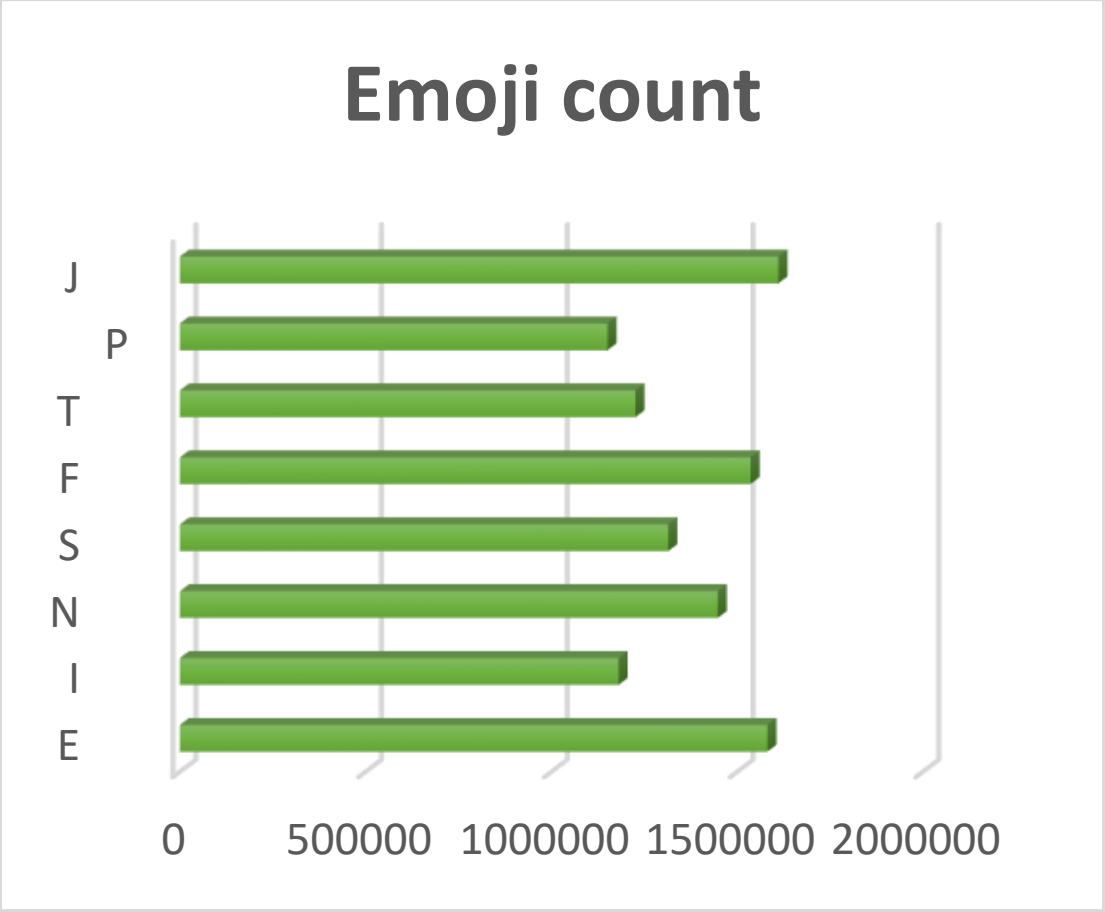

Figure 23: Emoji count per personality trait in Interactions

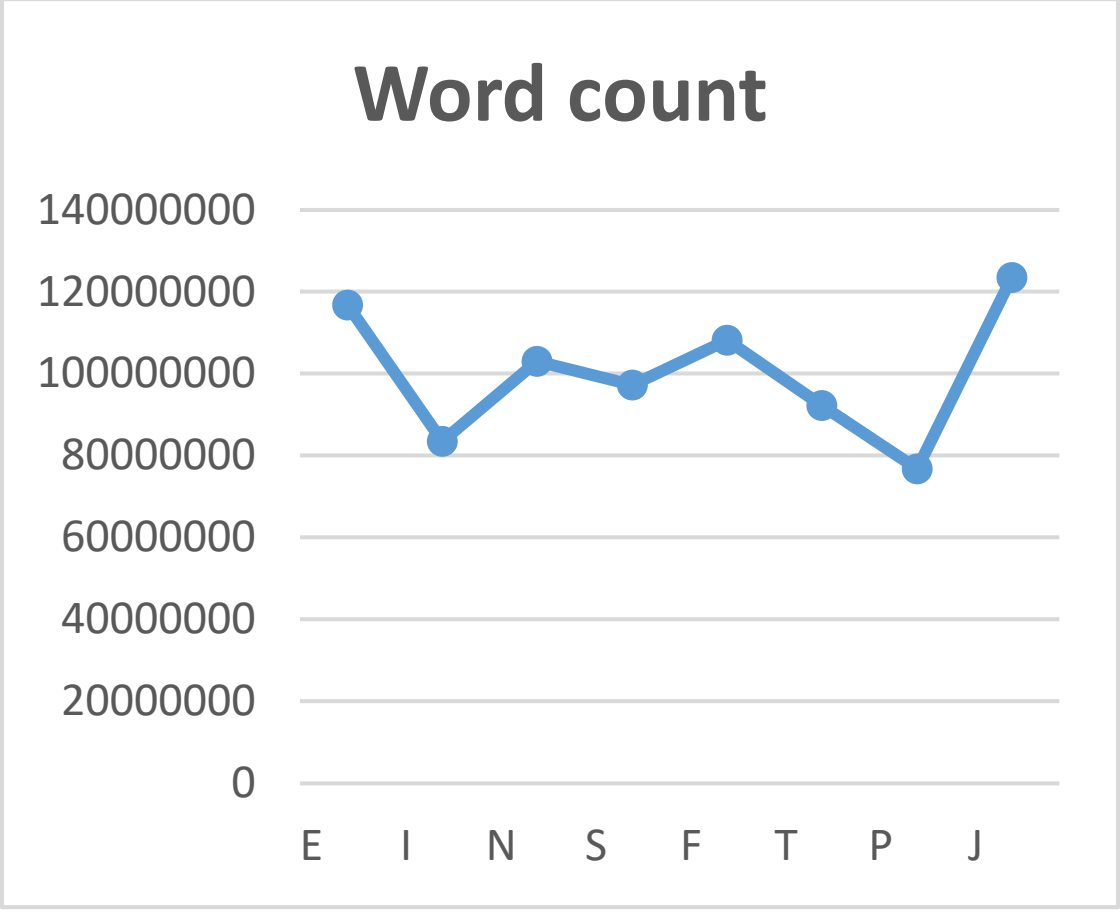

Figure 24: Word count per personality trait in Interactions

## 4.3 Personality Model:

We used the most frequent 1000 unigram, bigram, trigrams in our dataset to manage the feature set size while retaining valuable linguistic indicators.

Our goal was to predict personality traits based on various features, including Bag of Words, TF-IDF, TF-IDF n-gram, TF-IDF with n-gram on a character basis, and word2vector. For an insight into the embedding and considering the large nature of the dataset, we will include an illustrative sample for the bag of words: ('مختلف': 835, 'مال': 818) and for TF-IDF: (صورة: 0.6863265061584242, حلوة: 0.7272935631121525). As an example of word2vec output, the model yielded the following result when given the query (positive=['ملكة','سيدة'], negative=['رجل'], topn=1): [('عروس', 0.7768499255180359)].

We developed different prediction models using both individual and combined features. For combined features, we concatenate features within to assess model performance. The different machine learning algorithms tested initially were (fig 25):

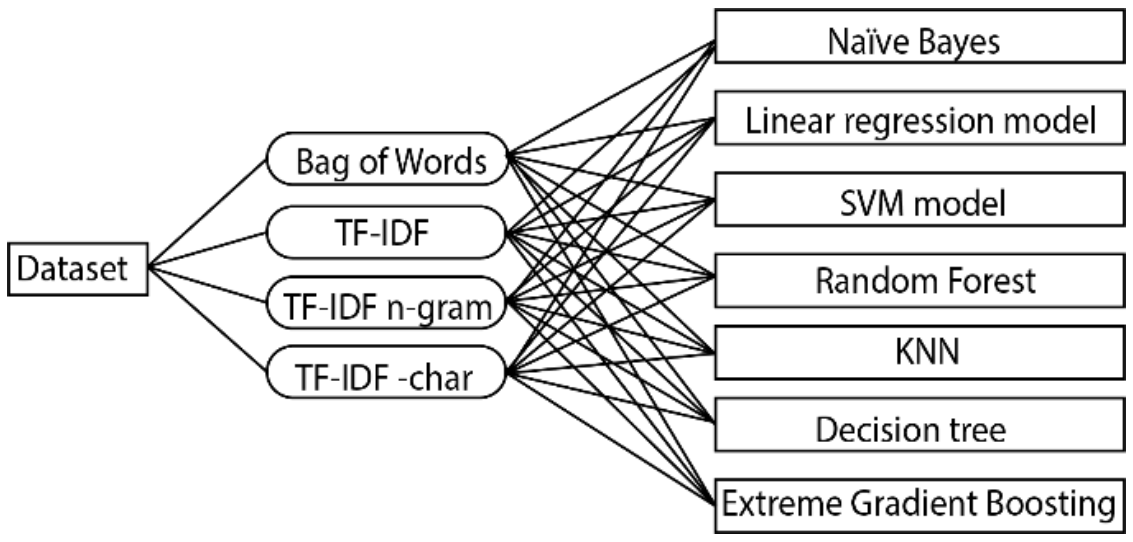

Figure 25: Applied machine learning algorithms

The performances reported in the following table:

Table 1: Performance of different machine learning methods with different features. Bold denotes the best result.

| Machine learning algorithm | BAG OF WORDS | | TF-IDF | | TF-IDF n-gram | | TF-IDF – char | |
|---|---|---|---|---|---|---|---|---|
| | Accuracy | F1 | Accuracy | F1 | Accuracy | F1 | Accuracy | F1 |
| Naïve Bayes | 24.9% | 24.34% | 23.94% | 22.64% | 20.84% | 19.56% | **28.73%** | **28.34%** |
| Linear regression model | 26.44% | 26.08% | 27.25% | 27.07% | 22.43% | 21.16% | **31.53%** | **31.37%** |
| SVM model | **17.36%** | **15.75%** | 15.94% | 11.42% | 14.52% | 13.2% | 16.46% | 16.17% |
| Random Forest | 31.45% | 30.33% | 33.54% | 33.2% | 22.54% | 21.33% | **38.79%** | **38.29%** |
| KNN | **25.53%** | **24.22%** | 25.7% | 25.3% | 15.83% | 14.47% | 16.37% | 16.02% |
| Decision tree | 31.38% | 30.47% | 32.11% | 31.7% | 22.58% | 21.36% | **37.84%** | **36.90%** |
| Extreme Gradient Boosting | 25.28% | 24.5% | 24.51% | 23.8% | 22.41% | 20.06% | **31.76%** | **31.38%** |

The performance of these models was unsatisfactory, leading us to explore neural networks for improved accuracy. We utilized a neural network with an input size of 1000, one hidden layer consisting of 100 neurons with a 'Relu' activation function, and an output layer with 16 neurons employing the 'softmax' activation function. The training commenced with a seed number of 42 for weight initialization, and a test size of 20% was allocated for validation. The optimizer, we employed was 'adam' with a learning rate of 0.001 and conducted training for 32 epochs. Given the nature of multi-class classification, we utilized 'sparse_categorical_crossentropy' as the loss function. For feed-forward deep learning, we replicated the aforementioned parameters but incorporated three hidden layers, each comprising 250 neurons. In the case of LSTM, we implemented 2 stacked LSTM layers. The obtained results are outlined below:

Table 2: Performance of different neural network methods with different features. Bold denotes the best result.

| Types | Methods | Accuracy |
|---|---|---|
| Neural Network | Bag of words and word2vec | 33.28% |
| | TF-IDF and word2vec | 33.61% |
| | TF-IDF (n-gram(2,3)) and word2vec | 23.44% |
| | TF-IDF(n-gram character base) and word2vec | **38.77%** |
| Feedforward Deep Learning | Bag of words and word2vec | **53.48%** |
| | TF-IDF and word2vec | 21.53% |
| | TF-IDF (n-gram(2,3)) and word2vec | 20.02% |
| | TF-IDF(n-gram character base) and word2vec | 15.19% |
| LSTM | Bag of words and word2vec | **26.40%** |
| | TF-IDF and word2vec | 18.74% |
| | TF-IDF (n-gram(2,3)) and word2vec | 20.47% |
| | TF-IDF(n-gram character base) and word2vec | 16.72% |

As indicated in (table.2) the combination of word2vector and bag of words features with feedforward deep learning model achieved the highest accuracy result of 53.48%. The experiments showed that using combined features improve the model performance, and removing word2vec significantly dropped the accuracy.

Finally, in order to concentrate more on the context within a sentence, knowing that the use of pre-trained models like BERT enhances the ability to capture complex contextual information, leading to more accurate and nuanced personality assessments, we worked on implementing the state-of-the-art method BERT [54].

Deep learning NLP models show key enhancements when trained on millions of annotated training samples, the thing that is not always available, to link the gap in data, researchers worked on creating methods for training general-purpose language illustration models using the unannotated text (pre-training). These general-purpose pre-trained models can then be fine-tuned on smaller task-specific datasets.

Pre-trained language illustrations can be either context-free or context-based. Context-based models generate a representation of each word based on the other words in the sentence.

BERT is based on the transformer model architecture, instead of LSTMs. A transformer works by performing a small, constant number of steps. In each step, it applies an attention mechanism to understand relationships between all words in a sentence, regardless of their respective position.

Key steps in using BERT:
-Pre-training:

1. Masked LM (MLM): Randomly masks words in a sentence and trains the model to predict these masked words.
2. Next sentence prediction (NSP): Trains the model to predict whether one sentence follows another.

-Fine-tuning:
Task-Specific Adjustment, after pre-training, BERT can be fine-tuned for specific tasks like text classification. For our task, this involved adding a classification layer on top of the pre-trained BERT model, as shown in (fig.26).

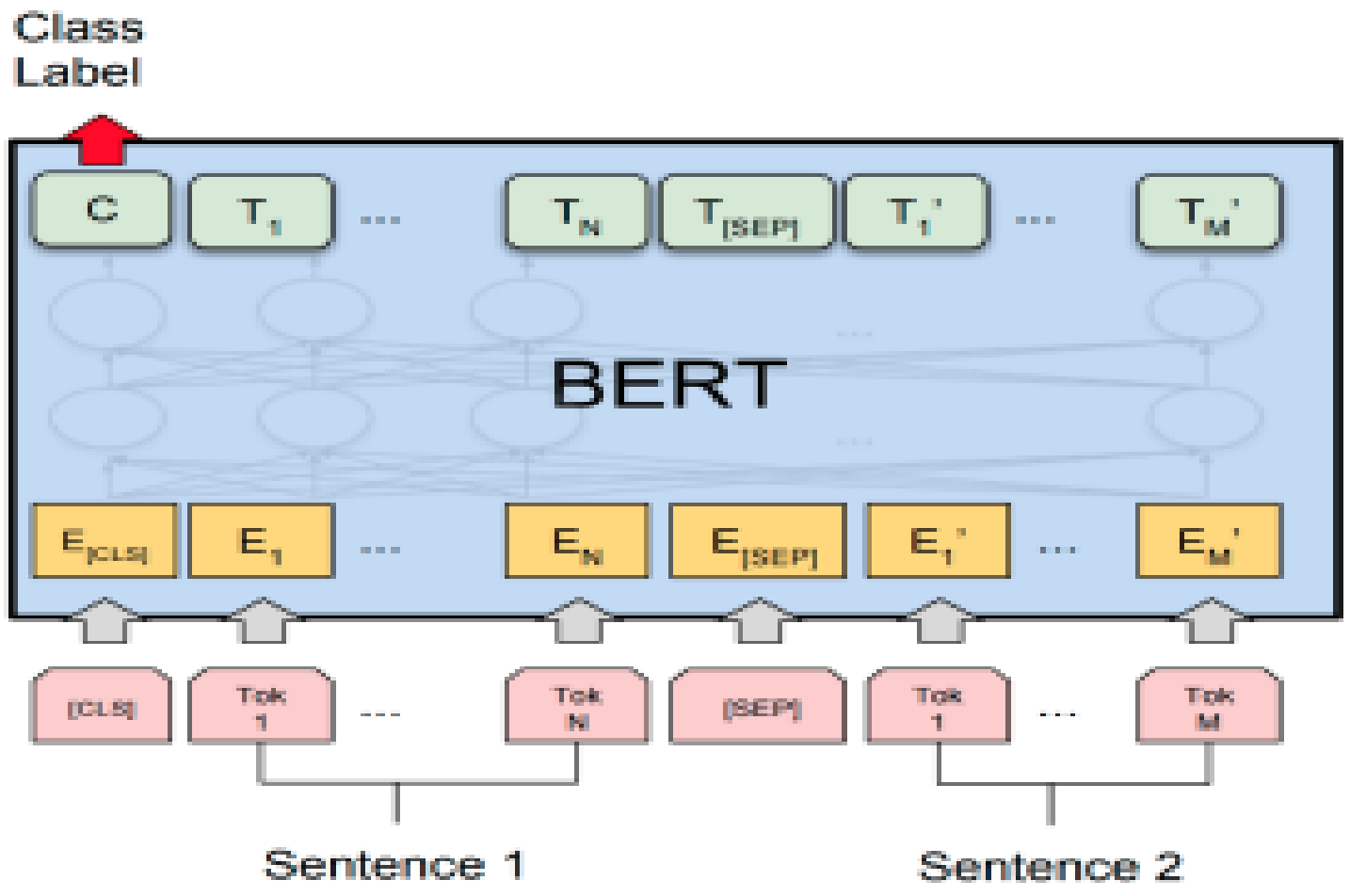

Figure 26: BERT fine-tuning [54]

We employed the general-purpose pre-trained BERT model for Arabic, AraBERT [55], utilizing the "bert-base-arabertv02" as the model.

Data preparation involved transforming personality labels into representative labels, and pre-processing was applied using the "ArabertPreprocessor." The pre-trained model and tokenizer were initialized, and the maximum sentence value length was set to 256.

The pre-trained model was configured for classification, incorporating a single classification layer. Evaluation metrics, specifically F-measure and accuracy, were employed. We utilized "AutoModelForSequenceClassification," loading "bert-base-arabertv02" as it provided a model with a sequence classification head.

The training duration spanned approximately 72 hours, with a test size of 20%. Our training loop comprised 16 epochs, featuring a learning rate of 2e-5, a batch size of 16, and a seed number of 42.

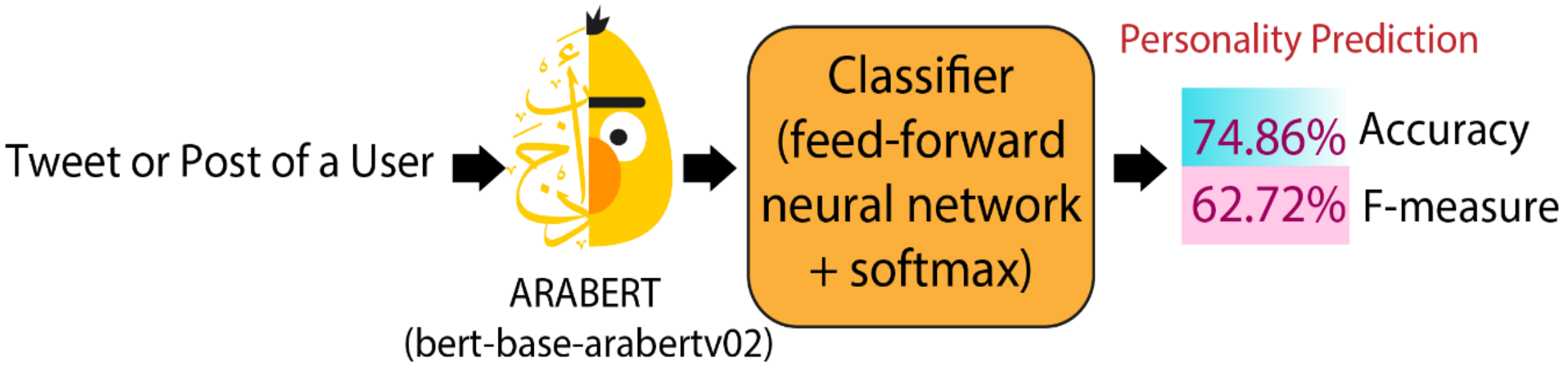

Figure 27: Results using ARABERT

As shown in the figure above (Fig.27), fine-tuning this pre-trained language model with our dataset results in an accuracy of 74.86%, surpassing the performance of other methods.

### 4.4 Personality effect on sentiment:

The sentiment analysis aimed to investigate how different personality types influence the sentiment expressed in tweets. By leveraging a balanced dataset, the "Balanced Corpus," which includes an equal number of tweets across various personality types, we used logistic regression model trained on a 14,744 tagged tweets dataset, to classify sentiment into positive, negative, and neutral categories. Subsequently, we analyzed the distribution of sentiment shares (positive, negative, and neutral) for each personality type, leading us to observe the following insights:

Most Positive Sentiments:
ESFJs (Consuls): Known for being “extraordinarily caring, social, community- minded people who are always eager to help” [53], ESFJs showed the highest proportion of positive sentiment. This aligns with their personality traits that are generally associated with positivity and social engagement.
ESTPs (Entrepreneurs): Characterized as “smart, energetic and very perceptive people, who truly enjoy living on the edge” [53], ESTPs also exhibited a high level of positive sentiment. Their enthusiasm for life and thrill-seeking nature likely contribute to this trend.

Most Negative Sentiments:
INFJ (Advocate): INFJs, who are "quiet and mystical, yet very inspiring and tireless idealists" [53], had a higher proportion of negative sentiment. Their introspective and often idealistic nature might lead to more critical or negative expressions in their tweets.
ISTJs (Logisticians): Known for being "practical and fact-minded individuals, whose reliability cannot be doubted" [53], ISTJs showed more negative sentiment. Their focus on reality and practicality might contribute to a more critical or cautious outlook.

Most Neutral Sentiments:
ESTJs (Executives): Known for being "excellent administrators, unsurpassed at managing things – or people" [53], ESTJs had the most neutral sentiment. Their pragmatic and organizational approach could contribute to a more balanced and less emotionally charged expression.
ENTJ (Commander): ENTJs, known as "bold, imaginative and strong-willed leaders, always finding a way – or making one" [53], also exhibited a significant amount of neutral sentiment. Their strategic and goal-oriented nature might result in more measured and less emotionally expressive communication.

This analysis highlights the relationship between personality traits and sentiment expression on social media. Understanding these patterns can help in tailoring communication strategies and improving engagement based on personality insights.

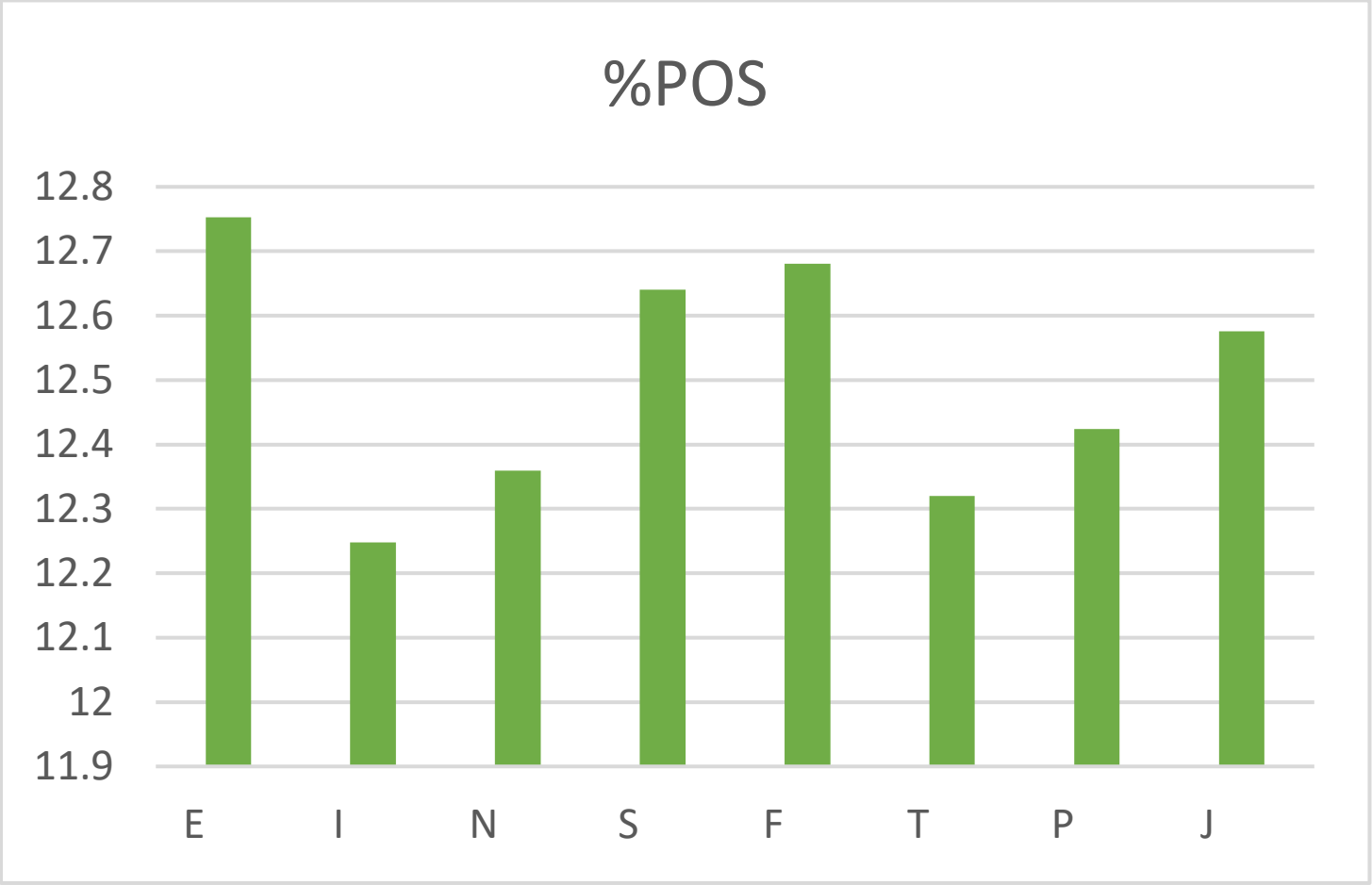

Figure 28: Positive Sentiment of each personality trait

As shown in (Fig.28) *extroverts* who are strengthened by spending time with people and tend to be more communicative and open, *sensors* who are interested in information they can straight see, hear, feel, and so on, and *feelers* who make decisions with their hearts and interested in how a choice will touch people have the most positive shares.

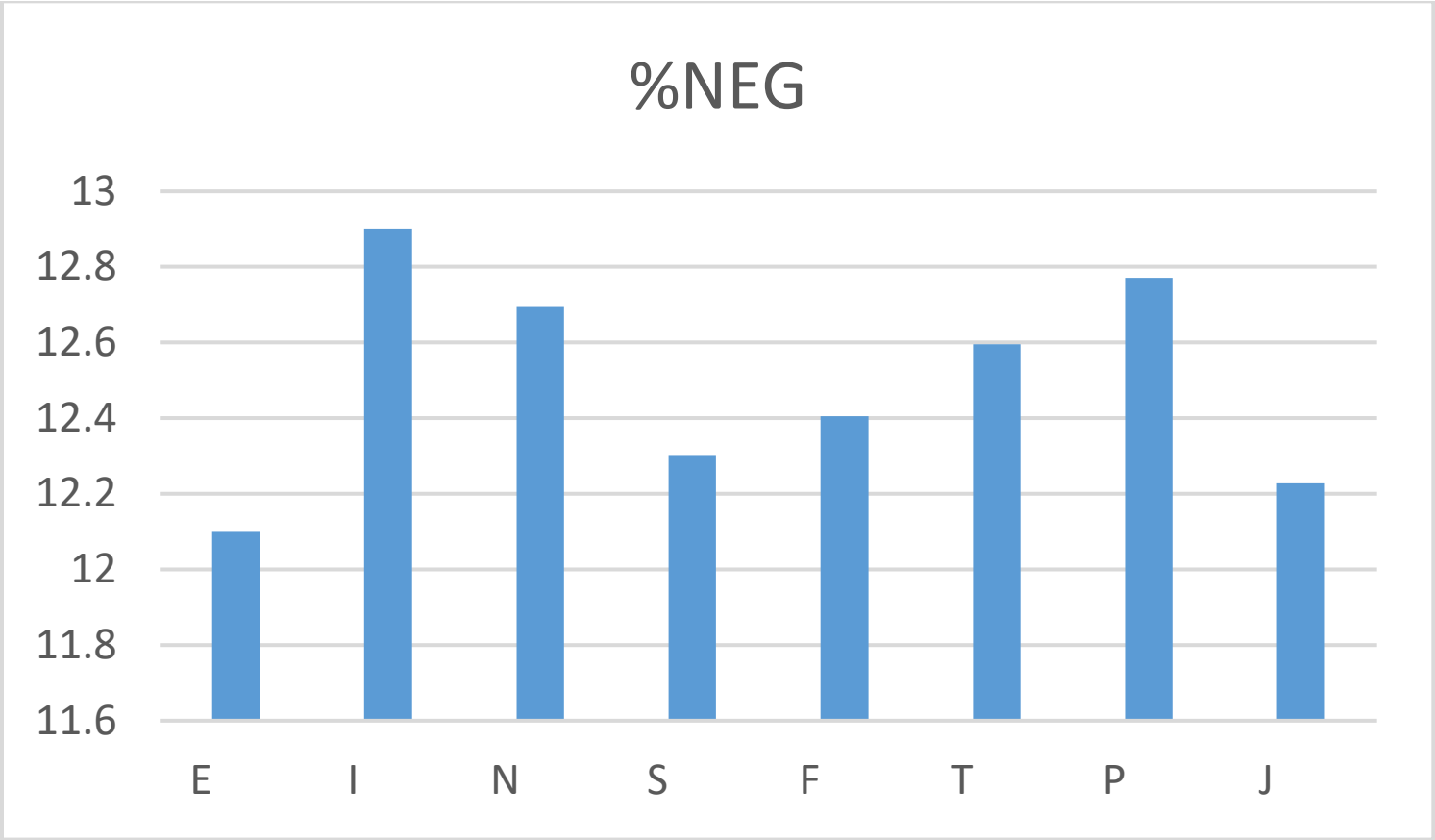

Figure 29: Negative Sentiment of each personality trait

*Introverts,* who love to spend quiet time alone or with a small group and tend to be more reserved, along with *Perceivers,* who like to leave things open to change their minds, and *Intuitives,* who focus on a more abstract level of thinking and more interested in theories, patterns, and explanations, have the most negative shares as shown in (Fig.29).

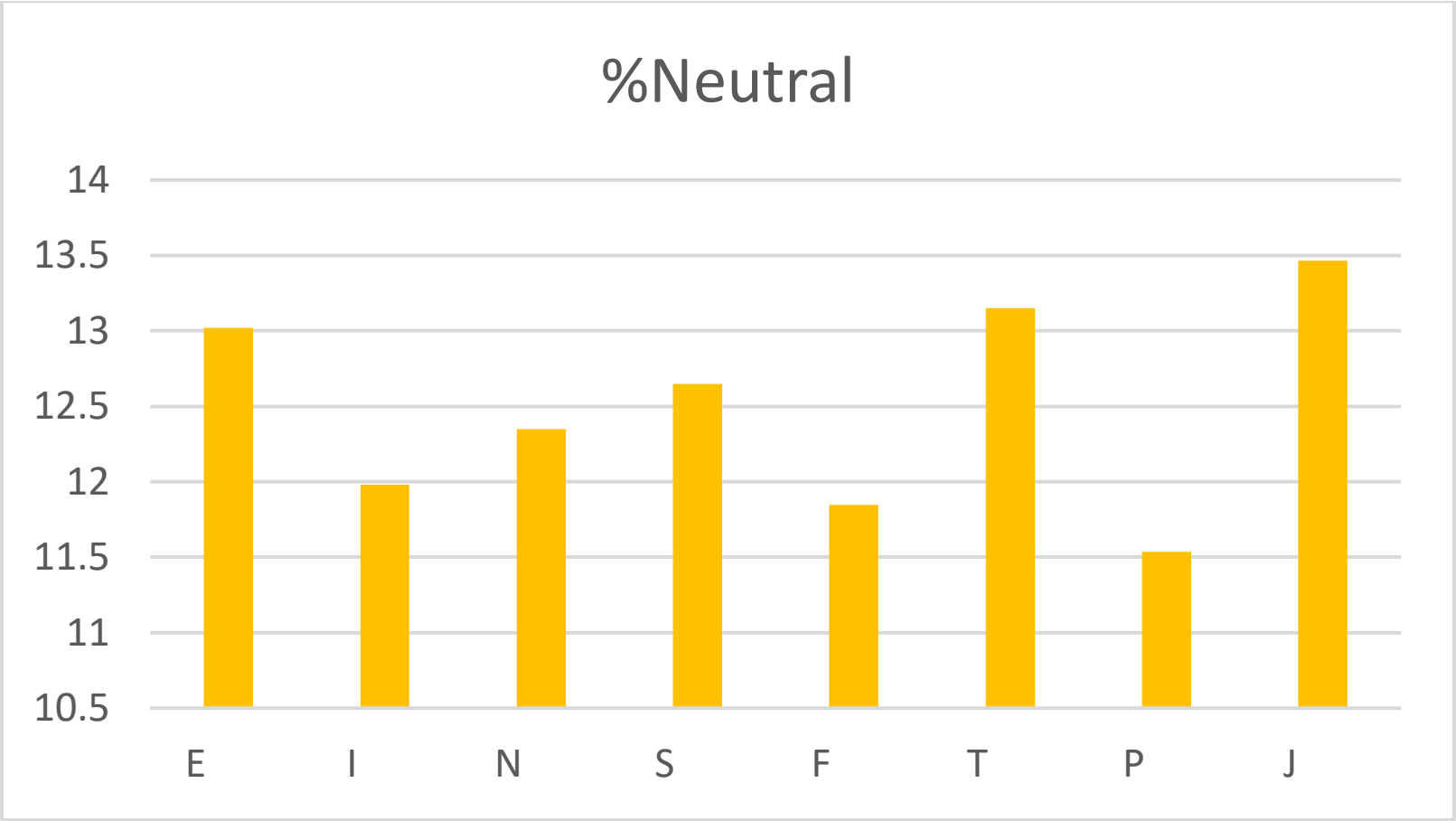

Figure 30: Neutral Sentiment of each personality trait

As shown in (Fig.30) *judgers* who appreciate structure and order and *thinkers* who tend to make choices with their heads have the most neutral shares.

The identified points are in harmony with the definition of each personality trait, offering logical reasoning for the obtained sentiment results. This underscores the direct relationship between personality and sentiment, providing a deeper understanding of how different personalities interact with their social media environment.

## 5 CONCLUSION

In this paper, we focused on extracting features for personality analysis, leveraging a substantial dataset we named it "AraPers". Notably, we improved the dataset's size compared to similar works, conducted a thorough analysis of the results, and correlated them with personality traits. Additionally, we explored the intricate connection between sentiment and personality, a relationship that became evident in our findings.

Conducting a comprehensive analysis of Arabic social media is imperative, especially given the growing number of Arabic language users on the web. Each facet we introduced serves to refine the analysis further, as demonstrated by the diverse results presented in this paper.

The proposed model holds practical applications, such as in recommender systems, customer relationship management (CRM), client acquisition and retention, cost-effective marketing, and understanding purchasing habits. Additionally, it can be valuable in targeting individuals during elections or influencing them.

NLP covers a vast subject area with numerous intersections with other scientific and humanities fields. Through continued experimentation, the relationship between human behavior and personality becomes clearer and more logically sound. However, it is essential to acknowledge the biases in our data collection process, particularly arising from the self-selection of participants who voluntarily disclosed their personalities online. Those engaging in such online activities may not represent the broader population. Furthermore, biases may exist due to cultural influences, geographic locations, and socioeconomic backgrounds. To enhance the external validity of future research, it is recommended to explore diverse sampling methods that address these biases. Despite these considerations, we believe our work contributes to a deeper understanding of the relationship between personality and social media interactions.